\documentclass[twoside,11pt]{article}

\usepackage{jmlr2e}
\usepackage{xcolor}
\usepackage{amsmath}
\usepackage{array}

\usepackage{algorithm, algpseudocode}
\usepackage[english]{babel}
\usepackage{amsmath}
\usepackage{amssymb}
\usepackage{graphicx}

\usepackage[caption=false]{subfig}

\def\EE{\mathbb{E}}

\newcommand{\KL}{\mathrm{KL}}

\newcommand{\MutI}{\mathrm{I}}
\newcommand{\Ent}{\mathrm{H}}

\newcommand{\qp}{\pi} %
\newcommand{\qq}{\pi} 
\newcommand{\pp}{\pi_0} 

\def\param{\theta}
\def\model{p_\param}
\def\varpost{q}
\def\varparam{\phi}
\def\latent{z}
\def\data{x}
\def\all{{1:N}}

\newcommand{\bp}[0]{\textit{behavior prior }}
\newcommand{\bps}[0]{\textit{behavior priors }}
\newcommand{\bpdot}[0]{\textit{behavior prior}}
\newcommand{\bpsdot}[0]{\textit{behavior priors}}

\ShortHeadings{Behavior Priors for Efficient Reinforcement Learning}{Tirumala et al.}
\firstpageno{1}

\begin{document}

\title{Behavior Priors for Efficient Reinforcement Learning}

\author{\name Dhruva Tirumala \email dhruvat@google.com \\
       \AND
       \name Alexandre Galashov \email agalashov@google.com \\
       \addr DeepMind, \\
       6 Pancras Square,
       Kings Cross, \\
       London, N1C 4AG       
       \AND
       \name Hyeonwoo Noh \email hyeonwoonoh@openai.com \\
       \addr OpenAI,
       3180 18th St, \\
       San Francisco, 
       CA 94110
       \AND       
       \name Leonard Hasenclever \email leonardh@google.com \\
       \AND
       \name Razvan Pascanu \email razp@google.com \\
       \AND
       \name Jonathan Schwarz \email schwarzjn@google.com \\
       \AND
       \name Guillaume Desjardins \email gdesjardins@google.com \\
       \AND
       \name Wojciech Marian Czarnecki \email lejlot@google.com \\
       \AND
       \name Arun Ahuja \email arahuja@google.com \\
       \AND
       \name Yee Whye Teh \email  ywteh@google.com \\
       \AND
       \name Nicolas Heess \email heess@google.com \\
       \addr DeepMind, \\
       6 Pancras Square, 
       Kings Cross, \\
       London, N1C 4AG
       }

\editor{Lorem and Ipsum}

\maketitle

\begin{abstract}
As we deploy reinforcement learning agents to solve increasingly challenging problems, methods that allow us to inject prior knowledge about the structure of the world and effective solution strategies becomes increasingly important.
In this work we consider how information and architectural constraints can be combined with ideas from the probabilistic modeling literature to learn \emph{behavior priors} that capture the common movement and interaction patterns that are shared across a set of related tasks or contexts. 
For example the day-to day behavior of humans comprises distinctive locomotion and manipulation patterns that recur across many different situations and goals.
We discuss how such behavior patterns can be captured using probabilistic trajectory models and how these can be integrated effectively into reinforcement learning schemes, e.g.\ to facilitate multi-task and transfer learning.
We then extend these ideas to latent variable models and consider a formulation to learn hierarchical priors that capture different aspects of the behavior in reusable modules. We discuss how such latent variable formulations connect to related work on hierarchical reinforcement learning (HRL) and mutual information and curiosity based objectives, 
thereby offering an alternative perspective on existing ideas. We demonstrate the effectiveness of our framework by applying it to a range of simulated continuous control domains, videos of which can be found at the following url: \url{https://sites.google.com/view/behavior-priors}.

\end{abstract}


\section{Introduction}
\label{sec:intro}

Recent advances have greatly improved data efficiency, scalability, and stability of reinforcement learning  (RL) algorithms leading to successful applications in a number of domains ~\citep{mnih2015human,silver2016mastering, heess2017emergence,riedmiller2018learning,andrychowicz2018learning,OpenAI_dota}. 
Many problems, however, remain challenging to solve or require large (often impractically so) numbers of interactions with the environment; a situation that is likely to get worse as we attempt to push the boundaries to tackle increasingly challenging and diverse problems. 

One way to address this issue is to leverage methods that can inject prior knowledge into the learning process. Knowledge extracted from experts or from previously solved tasks can help inform solutions to new ones, e.g. by accelerating learning or by constraining solutions to have useful properties (like smoothness). Accordingly there has been much interest in methods that facilitate transfer and generalization across different subfields of the RL community, including work on transfer learning \citep{rusu2016progressive, christiano2016transfer, teh2017distral, clavera2017policy, barreto2019transfer}, meta learning \citep{duan2016rl2, wang2016learning, finn2017modelagnostic, mishra2017simple, rakelly2019efficient, humplik2019meta} and hierarchical reinforcement learning (HRL) \citep{precup2000temporal, heess2016learning, bacon2017option, vezhnevets2017feudal, frans2018meta, wulfmeier2020compositional}. For example, recent success in the game of StarCraft \citep{vinyals2019grandmaster} relies on knowledge of useful skills and \emph{behaviors} that was extracted from expert human demonstrations. The ability to extract and reuse behaviors can also be leveraged in the multi-task setting. While solving several tasks simultaneously is nominally harder, the ability to share knowledge between tasks may in fact make the problem easier. For example, this is often the case when tasks form a curriculum where the solutions to easier problems can inform the solutions to harder ones \citep[e.g.][]{riedmiller2018learning}.

A related question that then arises naturally is which representations are best suited to capture and reuse prior knowledge. One approach is to directly use prior data as a way to constrain the space of solutions \citep{vinyals2019grandmaster, fujimoto2018offpolicy, wang2020critic}. An alternate approach that has gained much popularity expounds the use of hierarchical policies in order to combine together and sequence various skills and behaviors. These skills may be pre-defined, for instance, as motor primitives for control \citep{ijspeert2002learning, peters2009policy}, pre-learned with supervised learning or RL \citep[e.g.][]{heess2016learning,merel2018neural, paraschos2013probabilistic, lioutikov2017learning}, or can be learned on the fly through the use of sub-goal based architectures \citep{dayan1993feudal, vezhnevets2017feudal, nachum2018data}. Alternatively, they are often motivated as models better suited to represent temporally correlated behaviors \citep{sutton1999between, precup2000temporal, daniel2016probabilistic, bacon2017option, frans2018meta} and trained in an end-to-end manner. In this work, we present a unifying perspective to introduce \emph{priors} into the RL problem. The framework we develop presents an alternative view that allows us to understand some previous approaches in a new light.

Our approach views the problem of extracting reusable knowledge through the lens of probabilistic modeling. We build on the insight that policies combined with the environment dynamics define a distribution over trajectories. 
This perspective allows us to borrow tools and ideas from the rich literature on probabilistic models to express flexible inductive biases. We use this to develop a systematic framework around expressing prior knowledge, which we dub \bpsdot, and which can express knowledge about solutions to tasks at different levels of detail and generality.
They can be hand-defined or learned from data, integrated into reinforcement learning schemes, and deployed in different learning scenarios, e.g.\ to constrain the solution or to guide exploration. The framework admits for modular or hierarchical models which allow to selectively constrain or generalize certain aspects of the behavior such as low-level skills or high-level goals. The main contributions of our work can be summarized as follows:

\begin{itemize}
    \item \textbf{\textit{Behavior Priors} model trajectory distributions}: We develop the intuition of \bps as \textit{distributions over trajectories} that can be used to guide exploration and constrain the space of solutions. In this view, a good \emph{prior} is one that is general enough to capture the solutions to many tasks of interest while also being restrictive enough for tractable exploration.
    \item \textbf{Generalization and model structure}: 
    We demonstrate how the parametric form of the prior can be used to selectively model different aspects of a trajectory distribution, including simple properties such as smoothness but also complicated, long-horizon and goal-directed behavior. In particular we discuss how more restricted forms of the prior can encourage generalization and empirically show that this can lead to faster learning.
    \item \textbf{Hierarchy, modularity, and model structure}: We develop a general framework that supports learning \bp models that can be structured into multiple modules or hierarchical layers that communicate through latent variables. We show how such structure can be used to further control the inductive bias and to achieve a separation of concerns, e.g.\ of low-level skills and high-level goals, and how it can be used to selectively transfer or constrain aspects of the behavior.
    \item \textbf{Connections to Hierarchical RL}: We discuss the relationship between the proposed probabilistic trajectory models and other lines of work in the RL literature. 
    We show how common motifs from HRL can be motivated from the perspective of \bps, but also that model hierarchy is not a prerequisite for modeling hierarchically structured behavior.
    \item \textbf{Information theoretic regularization in RL}: We further highlight connections between our work and information theoretic regularization schemes applied in prior work. We find that our \bps can be motivated from the perspective of bounded rationality and information bottleneck, but that some of the models that we discuss also bear similarity to approaches motivated by curiosity and intrinsic motivation.
\end{itemize}

The rest of this work is split into two main parts. After some background for notation in Section \ref{sec:background}, in Sections \ref{sec:BP} and \ref{sec:experiments} we introduce our method and conduct an initial set of experiments for analysis and ablation to help ground these ideas. Following that, we extend our method to learn structured models in Section \ref{sec:structured_models} and empirically evaluate them in Section \ref{sec:hier_experiments}. In Section  \ref{sec:related}, we describe a unifying framework that places our work in a broader context of related work on HRL and methods based on mutual information and intrinsic rewards. Finally, we conclude with Section \ref{sec:conclusion}.
\section{Background}
\label{sec:background}
We now introduce some notation and background information that will serve as the basis of the work in later sections.
We start with some background definitions for RL and Markov decision processes (MDPs) \citep{suttonbarto}. For most of this work, we will limit our discussion to the discounted infinite-horizon setting for simplicity but our results also apply to the finite-horizon setting.

A Markov Decision Process (MDP) is defined by the following: $S$ and $A$ are state and action spaces, with $P: S \times A \times S \rightarrow \mathbb{R}_{+}$ a state-transition probability function or system dynamics and
$P_0: S \rightarrow \mathbb{R}_{+}$ an initial state distribution. We denote trajectories by $\tau = (s_0, a_0, s_1, a_1, \dots)$ and the state-action history at time step $t$ including $s_t$ (but not $a_t$) with $x_t = (s_0, a_0, \dots s_t)$.
We consider policies $\pi$ that are \emph{history-conditional} distributions over actions $\pi(a_t|x_t)$ \footnote{We generally work with history dependent policies since we will consider restricting access to state information from policies (for information asymmetry), which may render fully observed MDPs effectively partially observed.
}. 
Given the initial state distribution, transition dynamics and policy, the joint distribution over trajectories $\tau$ is given as:
\begin{align}
\pi(\tau) = 
    P_0(s_0)\prod_{t=0}^\infty P(s_{t+1}|s_{t},a_{t})\pi(a_t|x_t).
    \label{eq:trajectory_distribution:simple}
\end{align}
where $s_t\in S$ is the state at time $t\ge 0$ and $a_t\in A$ the corresponding action. 
For notational convenience, we have overloaded $\pi$ here to represent both the policy as well as the distribution over trajectories.
The learning objective is to maximize expected discounted returns, given by a reward function $r: S \times A \rightarrow \mathbb{R}$, and a discount factor $\gamma\in [0,1)$. 
Given a trajectory $\tau$, the discounted return is
\begin{align}
    R(\tau) = \sum_{t=0}^\infty \gamma^t r(s_t,a_t)
\end{align}
The expected discounted return over trajectories is then
\begin{align}
\mathcal{J}(\pi) = \EE_{\pi}[R(\tau)]
\label{eq:MDP_objective}
\end{align}
where the expectation is with respect to the trajectory distribution defined above. The goal of RL methods is to find an optimal policy $\pi^*(a | x)$ that maximizes the expected discounted return $\mathcal{J}(\pi)$ \citep{suttonbarto}.

Given a policy, the value functions $V^\pi(x)$ and $Q^\pi(x,a)$ are defined as the expected discounted return conditioned on history $x_t$ (and action $a_t$):
\begin{align}
V^\pi(x_t) &= \EE_{\pi}[R(\tau)|x_t]
= \EE_{\pi(a|x_t)}[Q^\pi(x_t,a)] \nonumber \\
Q^\pi(x_t,a_t) &= \EE_{\pi}[R(\tau)|x_t,a_t]
= r(s_t,a_t) + \gamma \EE_{P(s_{t+1}|s_t,a_t)}[V^\pi(x_{t+1})].
\end{align}
\section{Behavioral Priors for Control}
\label{sec:BP}

In this work the notion of trajectory distributions induced by policies (which we defined in Equation \ref{eq:trajectory_distribution:simple}) is an object of primary interest. For instance, we can think of 
the problem of finding a policy $\pi$ that maximizes Equation (\ref{eq:MDP_objective}) as that of finding a trajectory distribution for which the reward is maximal\footnote{Note that for MDPs and assuming no constraints on what a given policy class to which $\pi$ belongs can represent, a deterministic optimal policy will exist. If the transition dynamics are also deterministic then the trajectory distribution may collapse to a single trajectory which we can represent as a product of indicator functions.
}; and we can similarly think of different exploration strategies as inducing different trajectory distributions. This perspective of manipulating trajectory distributions allows us to bring intuitions from the probabilistic modeling literature to bear. In particular, we will introduce a method that allows us to express prior knowledge about solution trajectories in RL. 
Our starting point is the KL regularized objective
\citep{todorov2007linearly,kappen2012optimal,rawlik2012stochastic,schulman2017equivalence} \footnote{We derive this per-timestep KL-regularized objective in Appendix \ref{appendix:per_timestep_KL} when $\qp$ and $\pp$ have a form that is constrained by the system dynamics.}: 
\begin{align}
\mathcal{L}
&= \EE_\qp[ \sum_{t=0}^\infty \gamma^t r(s_t,a_t) - 
\gamma^t \KL[ \qp(a_t| x_t) || \pp(a_t| x_t) ]]. \label{eq:objective:generic_pertimestep}
\end{align}
where `KL' refers to the Kullback-Leibler divergence, a measure of similarity (or dissimilarity) between two distributions, which is defined as:
\begin{align}
\KL[ \qp(a_t| x_t) || \pp(a_t| x_t) ] &= \EE_{a_t \sim \qp(. | x_t)}[\log \frac{\qp(a_t | x_t)}{\pp(a_t | x_t)}] \nonumber
\end{align}

\subsection{KL-Regularized RL}
Intuitively, the objective in Equation (\ref{eq:objective:generic_pertimestep}) trades off maximizing returns with staying close (in the KL sense) to the trajectories associated with some reference behavior: $\pp$. Broadly speaking we can classify existing approaches that make use of this objective into two categories based on the choice of $\pp$. One approach, simply sets $\pp$ to be a uniform distribution over actions resulting in the entropy-regularized objective as used by ~\citet{ziebart2010modeling,schulman2017equivalence,haarnoja2017reinforcement,haarnoja2018soft,hausman2018learning}. This approach has been motivated in multiple ways \citep[e.g][]{ahmed2019understanding}, and, in practice, it has been shown to be helpful in preventing the policy from collapsing to a deterministic solution; improving exploration; and increasing the robustness of the policy to perturbations in the environment. 
A second class of algorithms optimize $\mathcal{L}$ with respect to both $\qp$ and $\pp$ and are often referred to as \emph{EM-policy search} algorithms. Examples of this style of algorithms include \citet{Peters10,toussaint2006probabilistic,rawlik2012stochastic,levine2013variational,levine2016end,montgomery2016guided,chebotar2016path,abdolmaleki2018maximum}.  
Although details vary, the common feature is that $\pp$ allows to implement a form of trust-region that limits the change in $\qp$ on each iteration but does not necessarily constrain the final solution. Different specific forms for $\qp$ and $\pp$ can then lead to alternate optimization schemes. 

In this work we focus on a different perspective. We consider cases where $\pp$ provides structured prior knowledge about solution trajectories. In this view $\pp$ can be thought of as a \bp, and Equation \eqref{eq:objective:generic_pertimestep} results in a trade off between reward and closeness of the policy $\qp$ to the prior distribution over trajectories defined by $\pp$. The prior $\pp$ can be seen as a constraint, regularizer, or shaping reward that guides the learning process and shapes the final solution. We discuss how $\pp$ can be learned from data; how the form of $\pp$ determines the kind of knowledge that it captures and how this in consequence determines how it affects the learning outcome in a number of different transfer scenarios.

\subsection{Multi-Task RL}
To gain more intuition for the objective in Equation (\ref{eq:objective:generic_pertimestep}) we first consider the multi-task RL scenario from \citet{teh2017distral}, with a distribution over tasks $p(w)$ where tasks are indexed by $w \in \mathcal{W}$. The tasks share the transition dynamics but differ in their reward functions $r_w$. We consider the KL-regularized objective with task-specific policy $\qp$ but a `shared' prior $\pi_0$ which has no knowledge of the task:
\begin{align}
\mathcal{L} = \sum_w p(w) \EE_{\qp_w} \left [ \sum_t \gamma^t r_w(s_t,a_t) - \gamma^t \KL[ \qp_w(a_t | x_t) || \pp(a_t | x_t) ] \right], 
\label{eq:ProbabilisticRL:MultitaskKL}
\end{align}
For a given $\pi_0$ and task $w$, we then obtain the optimal policy $\qp_w$ as follows: 
\begin{align}
\qp^*_w( a | x_t) =  \pp(a_t|x_t)\exp(Q^*_w(x_t,a) - V^*_w(x_t)),
\label{eq:ProbabilisticRL:OptimalExpert}
\end{align}
where $Q^*_w$ and $V^*_w$ are the optimal state-action value function and state value function  for task $w$ respectively which are given by,
\begin{align}
V^*_w(x_t) &= \max_{\pi_w \sim \Pi} \EE_{x_t \sim d_{\qp_{w,t}}} V^\pi_w(x_t) \nonumber \\
Q^*_w(x_t,a) &=  r(s_t, a) + \gamma \EE_{P(x_{t+1} | x_t, a)}[V^*_w(x_{t+1})] \nonumber
\end{align}
where $\Pi$ denotes the space of all policies and
\begin{align}
d_{\qp_{w,t}} &= P_0(s_0) \prod_{t^\prime=0}^{t-1} \qp_w(a_{t^\prime} | s_{t^\prime}) P(s_{t^\prime+1} |  s_{t^\prime}, a_t) \nonumber
\end{align}
On the other hand, given a set of task specific policies $\qp_w$, the optimal prior is given by:
\begin{align}
\pp^* &= \arg \min_{\pp} \sum_w p(w) \EE_{x_t \sim d_{\qp_{w,t}} } [ \KL [ \qp_w(\cdot | x_t) || \pp(\cdot | x_t) ]]
\label{eq:ProbabilisticRL:OptimalPriorRaw} \\
&= \sum_w p(w|x_t) \qp_{w}(a_t | x_t) \label{eq:ProbabilisticRL:OptimalPrior}
\end{align}

Together Equations (\ref{eq:ProbabilisticRL:OptimalExpert}) and (\ref{eq:ProbabilisticRL:OptimalPrior}) define an alternating optimization scheme that can be used to iteratively improve on the objective in Equation (\ref{eq:ProbabilisticRL:MultitaskKL}).
These results provide important intuition for the behavior of the KL regularized objective with respect to $\pi$ and $\pi_0$. 
In particular, Equation (\ref{eq:ProbabilisticRL:OptimalExpert}) suggests that given a prior $\pp$ the optimal task-specific policy $\qp_w$ is obtained by reweighting the prior behavior with a term that is proportional to the (exponentiated) soft-Q function associated with task $w$. Since the policy $\qp_w$ is the product of two potential functions ($\pp$, and $\exp Q$) it can effectively be seen as specializing the behavior suggested by the prior to the need of a particular task. Assuming that we can learn $Q^{*}_w$ we could directly use Equation (\ref{eq:ProbabilisticRL:OptimalExpert}) as a representation of the policy 
In practice, however, we normally learn a separately parametrized task specific policy $\pi_w$ as we detail in Section \ref{sec:BP:Algorithm}.

In contrast, the optimal prior $\pp^*$ for a set of task-specific experts $\pi_w$ is given as a \textit{weighted mixture} of these task specific policies where the weighting is given by the posterior probability 
of each of these task specific policies $\pi_w$ having produced trajectory $x_t$.
In other words, the optimal prior $\pp$ marginalizes over the task $w$ and produces the same trajectory distribution as if we first picked a task at random and then executed the associated expert. 
This is a useful policy insofar that, given an unknown task $w$, if we execute $\pp$ repeatedly, it will eventually execute the behavior of the associated expert $\qp_w$.
Since $\pp$ represents the behavior that is sensible to execute when the task is unknown it is also referred to as a \emph{default behavior} \citep{galashov2018information}.
In practice $p(w|x_t)$ is intractable to compute but we can easily sample from the expectation in Equation\ (\ref{eq:ProbabilisticRL:OptimalPriorRaw}) and fit a parametric model via distillation (i.e. by minimizing $\EE_{\qp_w} [ \log \pp(a_t | x_t)]$ as in Equation\ \ref{eq:ProbabilisticRL:OptimalPriorRaw}).

The optimal solutions for $\pp$ and $\qp_w$ thus satisfy the following basic intuition: For a given distribution over tasks $p(w)$ the optimal prior $\pp$ contains the optimal behavior for all tasks as a task-agnostic mixture distribution. The optimal task-specific policy $\qp_w$ then specializes this distribution to fit a particular task $w$. This result is a consequence of the properties of the KL divergence, in particular its asymmetry with respect to its arguments: For two distributions $q$ and $p$ the $\KL[q || p]$ will favor situations in which $q$ is fully contained in the support of $p$, and $p$ has support at least everywhere where $q$ has support. In other words, the KL divergence is mode-seeking with respect to $q$, but mode covering with respect to $p$ \citep[e.g.][]{bishop2006pattern}. In the next section we will discuss how these properties can be exploited to control generalization in RL.

\subsection{General Information Asymmetry for \textit{Behavioral Priors}}

Studying Equation (\ref{eq:objective:generic_pertimestep})
we notice that if $\pp$ had access to all the information of $\qp$ 
, then the optimal solution would be to just set $\pp^*=\qp^*$. Thus, it is the fact that we constrain the processing capacity of $\pp$, 
by removing $w$ from the information that $\pp$ has access to
that results in a default policy that \emph{generalizes} across tasks in Equation (\ref{eq:ProbabilisticRL:OptimalPrior}). 
We can extend this intuition by considering  priors $\pp$ which are restricted by limiting their modeling capacity or the information they have access to more generally. To make this precise we split $x_t$ into two disjoint subsets $x_t^G$ and $x_t^D$ \footnote{In line with terminology from the previous section we use superscripts $D$ for `default' and $G$ for `goal' although we now consider a more general use.
} and allow $\pp$ access only to $x_t^D$, i.e.\ $\pp(\cdot| x_t) = \pp(\cdot| x_t^D)$ while $\qp$ retains access to all information $\qp(\cdot| x_t) = \qp(\cdot| x_t^G,x_t^D)$.
The objective in Equation (\ref{eq:objective:generic_pertimestep}) then becomes:
\begin{align}
\mathcal{L}
&= \EE_\qp[ \sum_t \gamma^t r(s_t,a_t)] - 
\gamma^t \KL[ \qp(a_t| x_t) || \pp(a_t| x^D_t) ]. \label{eq:objective:info_asym}
\end{align}

In the notation of the previous section $x_t^G = w$ and $x_t^D = x_t$. But we could, for instance, also choose $x_t^D = (a_0, a_1, \dots a_t)$ and thus allow $\pp$ to only model temporal correlations between actions. In our experiments we will consider more complex examples such as the case of a simulated legged `robot' which needs to navigate to targets and manipulate objects in its environment. We will give $\pp$ access to different subsets of features of the state space, such as observations containing proprioceptive information (joint positions, velocities, etc.), or a subset of exteroceptive observations such as the location and pose of objects, and we study how the behavior modeled by $\pp$ changes for such choices of $x_t^D$ and how this affects the overall learning dynamics. 

The interplay between $\qp$ and $\pp$ is complex. An informative $\pp$ simplifies the learning problem since it effectively restricts the trajectory space that needs to be searched\footnote{Concretely, the second term in Equation (\ref{eq:ProbabilisticRL:MultitaskKL}) can be considered as a shaping reward; and Equation\ (\ref{eq:ProbabilisticRL:OptimalExpert}) emphasizes how $\pp$ restricts the solutions for $\qp$. Trajectories not included in the support of $\pp$ will not be included in the support of $\qp$ and are thus never sampled.}. In the multi-task scenario  of the previous section, and in the more general case of asymmetry described in this section, $\pp$ can generalize shared behaviors across tasks (or across different values of $x_t^G$, i.e.\ across different parts of the trajectory space). However, the presence for $\pp$ will also affect the solution learned for $\qp$ in the sense that the overall objective will favor $\qp$ that are closer to $\pp$, for instance because they have little reliance on $x_t^G$. This may improve robustness of $\qp$ and help generalization to settings of $x_t^G$ not seen during training. We will discuss this last point in more detail in Section \ref{sec:related} when we discuss the relation to information bottleneck.

Finally, it is worth noting that when, as is usually the case, $\pp$ is separately parametrized as $\pp^\phi$ and learned (e.g.\ as described in Section \ref{sec:BP:Algorithm}), then its parametric form will further affect the learning outcome. For instance, choosing $\pp^\phi$ to be a linear function of $x_t$\footnote{
When we say that $\pp$ (or $\qp$) is a linear function of $x_t$ we mean that the parameters of the policy distribution, e.g.\ the mean and the variance in if $\pp$ is Gaussian, are linear functions of $x_t$.
} will limit its ability to model  the dependence of $a_t$ on $x_t$. Furthermore, choosing the parametric model for $\pp^\phi$ to be a Gaussian may limit its ability to model the mixture distribution in Equation (\ref{eq:ProbabilisticRL:OptimalPrior}). Here, too, the direction of the $\KL$ will force $\pp^{\phi}$ to extrapolate. Overall, the information that $\pp$ is conditioned on, the parametric form of the function approximator, and the form of the sampling distribution determine the behavior model that $\pp$ learns and how it encourages generalization across the trajectory space.

Another interesting aspect of the joint optimization scheme is that as the prior learns, it can play the role of a `shaping reward'. The structure that it learns from the solutions to some tasks can be used to \textit{inform} solutions to others. For instance, in a locomotion task, the same underlying movement patterns are needed for goals that are close by and those that are further away. We will explore some of these effects in our experiments in Section \ref{sec:experiments} and include a simple 2-D example to help develop an intuition for this in Appendix \ref{appendix:additional_results}.

\subsection{Connection to variational Expectation Maximization (EM)}
The entropy regularized objective is sometimes motivated by comparing it to the problem of computing a variational approximation to the log-marginal likelihood (or, equivalently, the log-normalization constant) in a probabilistic model. Similarly, we can establish a connection between Equation 
(\ref{eq:ProbabilisticRL:MultitaskKL}) and the general variational framework for learning latent variable models \citep{dempster1977maximum} (which we briefly review in Appendix \ref{appendix:probablistic_background}). 
In that setup the goal is to learn a probabilistic model $\model$ of some data  $\data_\all = ( \data_1, \dots \data_N )$ by maximizing the log marginal likelihood $\log \model(\data) = \sum_i \log \model(\data_i)$ where $\model(\data) = \int \model(\data, \latent)d\latent$. This likelihood can be lower bounded
by  $\sum_i \EE_\varpost(\latent | \data_i)[\log \model(\data_i| \latent ) - \log(\frac{\varpost_\phi(\latent | \data_i)}{\model(\latent)})]$ where $\varpost_\phi$ is a learned approximation to the true posterior. 

We can draw a correspondence between this framework and the objective from Equation (\ref{eq:objective:generic_pertimestep}). First consider each data point $\data_{i}$ is a task in a multi-task setting where each latent $z$ defines a trajectory $\tau$ in the corresponding MDP. In other words the conditional probability $\log \model(\data_i| z)$ measures the sum of rewards generated by the trajectory $z$ for task $i$. Note that in this case $\model$ has no learnable parameters and is a measure of the goodness of fit of a trajectory for a given task. The \emph{prior} $\model(\latent)$ is now a distribution over trajectories generated under the \bp and system dynamics. Similarly the posterior $\varpost_\phi(\latent | \data_i)$ defines a distribution over trajectories under the \emph{policy} and system dynamics. We will sometimes refer to the policy as the `posterior' because of this equivalence. 

\begin{algorithm*}[tb]
  \caption{Learning \textit{priors}: SVG(0) with experience replay}
  \label{alg:rs0_distral}
\begin{algorithmic}
  \scriptsize
  \State Policy: $\pi_\theta(a_t|x_t)$ with parameters $\theta$
  \State {\color{blue} \textit{Behavioral Prior}: $\pi_{0,\phi}(a_t|x^D_t)$}
  \State Q-function: $Q_\psi(a_t, x_t)$ with parameters $\psi$
  \\
  \State Initialize target parameters $\theta^\prime \leftarrow \theta$,
  \hspace{0.2cm} {\color{blue} $\phi^\prime \leftarrow \phi$}, \hspace{0.2cm} $\psi^\prime \leftarrow \psi$.
  \State Hyperparameters: {\color{blue} $\alpha$}, $\alpha_H$, $\beta_{\qp}$, {\color{blue}$\beta_{\pp}$}, $\beta_{Q}$
  \State Target update counter: $c \leftarrow 0$
  \State Target update period: $P$
  \State Replay buffer: $\mathcal{B}$
  \\
  \For{$j=0, 1, 2, ... $}
  \State Sample partial trajectory $\tau_{t:t+K}$ generated by behavior policy $\mu$ from replay $\mathcal{B}$:\\ \hspace{1.6cm}$\tau_{t:t+K}=(s_t,a_t,r_t,...,r_{t+K})$
  \\
  \For{$t^\prime=t, ...t+K$}\\
  \hspace{1.6cm}{\color{blue} $\hat{\KL}_{t^\prime}=\KL\left[\pi_\theta(.|x_{t^\prime})\|\pi_{0,{\phi^\prime}}(.|x^D_{t^\prime})\right]$}
  \Comment{Compute KL}\\
  \hspace{1.6cm}{$\color{blue}\hat{\KL}^\mathcal{D}_{t^\prime}=\KL\left[\pi_\theta(.|x_{t^\prime})\|\pi_{0,{\phi}}(.|x^D_{t^\prime})\right]$}
  \Comment{Compute KL for distillation}\\
  \hspace{1.6cm}$\hat{\text{H}}_{t^\prime}=\EE_{\pi_\theta(a|x_{t^\prime})}[\log\pi_\theta(a|x_{t^\prime})]$  
  \Comment{Compute action entropy}\\
  \\
  \hspace{1.6cm}$\hat{V}_{t^\prime}
        =\EE_{\pi_\theta(a|x_{t^\prime})}
        \left[Q_{\psi^\prime}(a,x_{t^\prime})\right]-{\color{blue}\alpha\hat{\KL}_{t^\prime}}$
  \Comment{Estimate bootstrap value}\\
  \hspace{1.6cm}$\hat{c}_{t^\prime}=\lambda \min\left(\frac{\pi_\theta(a_{t^\prime}| x_{t^\prime})}{\mu(a_{t^\prime}| x_{t^\prime})}, 1\right)$
  \Comment{Estimate traces~\citet{munos2016safe}}\\  
  \hspace{1.6cm}
  $\hat{Q}^R_{t^\prime}=Q_{\psi^\prime}(a_{t^\prime},x_{t^\prime})
        +
        \sum_{s \ge t^\prime}\gamma^{s-t^\prime}\left(\prod_{i=t^\prime}^{s}\hat{c}_i\right)
        \left(r_s+\gamma \hat{V}_{s+1} - Q_{\psi^\prime}(a_s,x_s)\right)$\\
  \Comment{Apply Retrace to estimate Q targets}\\
  \EndFor \\
  \\
  \State
  $\hat{L}_Q=\sum_{i=t}^{t+K-1}\|\hat{Q}^R_i-Q_\psi(a,x_{i})\|^2$
  \Comment{Q-value loss}
  \\
  \State
  $\hat{L}_\pi=\sum_{i=t}^{t+K-1}\EE_{\pi_\theta(a|x_{i}, \eta)}Q_{\psi^\prime}(a,x_{i})-{\color{blue} \alpha\hat{\KL}_i} + \alpha_\text{H}\hat{\text{H}}_i$  
  \Comment{Policy loss} 
  \\
  \State
  {\color{blue} $\hat{L}_{\pi_0}=\sum_{i=t}^{t+K-1}\hat{\KL}_i^\mathcal{D}$}
  \Comment\textit{Behavior Prior} loss\\
  \State
  $\theta \leftarrow \theta+\beta_\qp\nabla_\theta\hat{L}_\pi$ \hspace{0.3cm}
  {\color{blue} $\phi \leftarrow \phi+\beta_{\pp}\nabla_\phi\hat{L}_{\pi_0}$}
  \hspace{0.3cm}
  $\psi \leftarrow \psi-\beta_{Q}\nabla_\psi\hat{L}_{Q}$
  \State Increment counter $c \leftarrow c+1$
  \If{$c > P$}
  \State Update target parameters $\theta^\prime \leftarrow \theta$,
  \hspace{0.2cm} {\color{blue} $\phi^\prime \leftarrow \phi$}, \hspace{0.2cm} $\psi^\prime \leftarrow \psi$
  \State $c \leftarrow 0$
  \EndIf
  \EndFor
\end{algorithmic}
\end{algorithm*}

\subsection{Algorithmic considerations}
\label{sec:BP:Algorithm}
The objective for training \bps from Equation (\ref{eq:objective:info_asym}) involves solving a regularized RL problem similar to \citet{schulman2017equivalence,hausman2018learning,haarnoja2018soft}, as well as an additional distillation step for optimizing the prior. For this work, we adapt the off-policy actor-critic algorithm SVG-0 of \citet{heess2015learning} although as such our method can easily be incorporated into any RL algorithm. The SVG-0 algorithm proceeds by collecting trajectories from some behavior policy $\mu$ which are stored into a replay buffer \citep[as in][]{mnih2015human}. $K$ length tuples of experience data $(s_t, a_t, r_t, ..., s_{t+K}, a_{t+K}
, r_{t+K})$ are then sampled from the buffer for learning. This data is used to train 3 components: a critic; an actor and a \bpdot. Additionally, we use target networks for each of these components as in  \cite{mnih2015human} which we found stabilized learning.
\paragraph{Critic update:}
In order to estimate the state-action value function $Q(s, a)$, we use the retrace estimator \citep{munos2016safe} which has been shown to reduce variance in the multistep off-policy setting. For some timestep $t$ the estimate is given by:
\begin{align}
  Q^R(x_{t}, a_{t}) &= Q(x_{t}, a_{t})
        +
        \sum_{s \ge t}\gamma^{s-t}\left(\prod_{i=t}^{s}c_i\right)
        \left(r_s+\gamma V(x_{s+1}) - Q(x_s, a_s)\right)  \label{eq:retrace} \\
 c_{i}&=\lambda \min\left(\frac{\pi(a_{i}| x_{i})}{\mu(a_{i}| x_{i})}, 1\right) \nonumber
\end{align}        
where $r$ is the task reward, $\pi$ is the current policy, $c_i$ represents a form of clipped importance sampling for off-policy correction and $V$ is given by:
\begin{align}
     V(x_t) &= \EE_{a \sim \qp(.| x_t)}\left[Q(x_t, a)\right] - \alpha \KL\left[\qp(.|x_t)\|\pp(.|x^D_t)\right] \nonumber
\end{align}
Note that we incorporate the $\KL$ term from Equation (\ref{eq:objective:info_asym}) for timestep $s > t$ through the bootstrap value $V$ in Equation (\ref{eq:retrace}). This is because for the tuple $(x_{t}, a_{t}, r_{t})$ at time $t$, the reward $r_{t}$ is the effect of the \emph{previous} action $a_{t-1}$ while $\KL\left(\qp(a_{t}|x_{t}) || \pp(a_{t}|x_{t})\right)$ corresponds to the \emph{current} action $a_{t}$. The KL for timestep $t$ is thus added to the policy update instead.
With this, we can write down the objective for the critic as follows:
\begin{align}
{L}_Q&=\sum_{i=t}^{t+K-1}\|Q^R(x_i, a_i) - Q(x_{i}, a_i)\|^2 \nonumber
\end{align}

\paragraph{Policy update:} The SVG-0 algorithm trains a stochastic policy from $Q(s, a)$ value function via reparameterization of the gradient. That is:
\begin{align}
\nabla_\theta \EE_{\qp_\theta} \left[ Q(s, a) \right] &= \EE_{\rho(\eta)} \left [\nabla_a Q(s, a) \nabla_\theta \qp_\theta (a| s, \eta) \right] \nonumber
\end{align}
where $\eta$ is random noise drawn from a distribution $\rho$. As described above, we include an additional $\KL$ term to the policy loss for the objective in Equation (\ref{eq:objective:info_asym}). We also include an entropy bonus term which we found improves performance on some domains. Note that this entropy objective is also used for the baseline in our experiments in Sections \ref{sec:experiments} and \ref{sec:hier_experiments}. The resulting entropy-regularized objective is thus also effectively the same as the one considered by \cite{haarnoja2018soft}.
Putting these together we can write down the objective for the policy as:
\begin{align}
L_\pi &= \sum_{i=t}^{t+K-1}\EE_{a \sim \pi(.|x_i, \eta)}Q(x_{i}, a) - \alpha \KL\left[\qp(.|x_i) |\pp(.|x^D_i)\right] + \alpha_{\Ent}{{\Ent_i(\qp)}}  \nonumber
\end{align}
where the entropy $\Ent_t$ is given by
\begin{align}
\Ent_{t}(\qp)=\EE_{a \sim \qp(.|x_t)}[\log\qp(a|x_t)] \nonumber
\end{align}
\paragraph{Prior update:} Finally we train the \bp $\pp(.| x^D_t)$ to match the policy distribution $\qp(.| x_t)$ with the following objective:
\begin{align}
    L_{\pp} & =\sum_{i=t}^{t+K-1} \KL\left[\qp(.|x_i) \|\pp(.|{x^D}_i)\right] \nonumber
\end{align}
We refer to this form of distribution matching through minimizing the $\KL$ as `distillation' in keeping with \citet{teh2017distral}. The full procedure used for training is shown in Algorithm \ref{alg:rs0_distral}, where the additions for learning a \emph{prior} are highlighted in blue. We used separate ADAM optimizers \citep{kingma2014adam} for training the critic, policy and \bpdot. A full list of hyperparameters used is presented in Appendix \ref{appendix:experiment_details}.
\section{Experiments}
\label{sec:experiments}
In this section, we analyze the effect of \bps experimentally on a number of simulated motor control domains using walkers from the DeepMind control suite \citep{tassa2018control} developed using the MuJoCo physics engine \citep{todorov2012mujoco}. 
The purpose of these experiments is to understand how \textit{priors} with various capacity and information constraints can learn to capture general task-agnostic behaviors at different levels of abstraction, including both basic low-level motor skills as well as goal-directed and other temporally extended behavior.
We consider a range of multi-task and transfer problems with overlapping solution spaces that we can model at varying levels of detail and generality. 

Our range of `Locomotion' tasks
requires a simulated agent to reach various goal locations or manipulate objects in its environment. The locations of the goals, objects and walker are chosen at random for every episode. In other words, each task is a distribution over goals and targets, some of which are harder to solve than others. However all of them share a common \emph{underlying structure}; their solutions involve consistent patterns of interaction between the agent and environment. For instance, a common element underpinning these tasks is that the gait or underlying pattern of movement of any agent that solves it is largely goal independent. More specific and less general behavior that recurs across tasks includes goal-directed locomotion or object interactions. 

In Section \ref{sec:info_asym_base} we show information constraints with simple Multilayer Perceptron (MLP) models can learn to model these behaviors. For instance, we demonstrate that \textit{priors} without any hierarchical structure can learn to model temporally correlated patterns of movement for a complex 21 degree-of-freedom (DoF) humanoid body. 
Apart from information constraints, we are also interested in understanding how architectural choices can affect the kinds of behaviors modeled under the prior. With this in mind, we include problems that are specifically designed to exhibit structure at different temporal scales: In the `Sequential Navigation' tasks an agent must visit several targets in \emph{sequence}, with some target sequences being much more likely to occur. In this setting we show how some architectures can learn to model these dynamics.

We vary the complexity of these tasks using more complicated bodies (\textit{Locomotion (Humanoid)}); changing the number of boxes or targets (\textit{Manipulation (2 boxes, 2 targets)}); or by combining different tasks (\textit{Locomotion and Manipulation}).
All the tasks that we consider in this work use sparse rewards. Sparse reward tasks are typically harder to learn but are easier to specify since they do not require hand engineered per-timestep rewards. This setting provides an interesting context in which to study \bpsdot. As the \textit{prior} learns to model solutions for some instances, it can help \emph{shape} the learning process and guide the policy to discover solutions to new ones. As we show below, this can lead to faster convergence for the policy on a range of tasks.

For convenience, we have summarized the tasks used in this section and Section \ref{sec:hier_experiments} in Table \ref{table:task_list}. Videos of all tasks can be found on our accompanying website: \url{https://sites.google.com/view/behavior-priors}. 

Unless otherwise specified, learning curves show returns (on the Y-axis) as a function of the number of environment steps processed by the learner (X-axis) and for each curve, we plot the mean of the best performing configuration averaged across 5 seeds with shaded areas showing the standard error. A complete list of hyperparameters considered is included in Appendix \ref{appendix:experiment_details}. All experiments were run in a distributed setup using a replay buffer with 32 CPU actors and 1 CPU learner.
\begin{table}[t]
\scriptsize
\centering
\begin{tabular}{ | m{2.3cm} | m{2.5cm}| m{1.6cm} |  m{1.2cm} |  m{2cm} | m{2cm} |} 
 \hline
 \textbf{Task Name} & \textbf{Description} & \textbf{Body} & \textbf{Action Dim.} & \textbf{Obs.} & \textbf{Obs. to \textit{Prior}} \\
 \hline\hline
 
 \textit{Navigation Easy} & Visit targets in a sequence where the sequence is history dependent. & PointMass / Ant & 2/8 & Proprio + Targets + next target index & Proprio + targets \\ 
 \hline
 
 \textit{Navigation Hard} & Visit targets in order. & PointMass / Ant & 2/8 & Proprio + Targets & Proprio + Targets \\ 
 \hline  
 
 \textit{Locomotion (Humanoid)} & Follow a moving target  & Humanoid & 23 & Proprio +
  Target location & Varies (See text) \\ 
 
 \hline
 
 \textit{Locomotion (Ant)} & Go to one of 3 targets  & Ant & 8 & Proprio + Targets + Index of goal & Varies (See text) \\ 

 \hline
 
 \textit{Locomotion (Gap)*} & Move forward and jump across a variable length gap. & Ant & 8 & Proprio + Global position + Gap length & Proprio \\ 

 \hline
 
 \textit{Locomotion and Manipulation} & Move a box to a specified target and then move to another target.  & Ant & 8 & Proprio + Box + Targets + Task encoding & Varies (See text) \\ 
 
 \hline
 
 \textit{Manipulation (1 box, 1 target)*} & Move a box to a target location.  & Ant & 8 & Proprio + Box + Targets + Task encoding & Proprio + Box \\
 
 \hline
 
 \textit{Manipulation (1 box, 3 target)*} & Move a box to one of 3 target locations.  & Ant & 8 & Proprio + Box + Targets + Task encoding & Proprio + Box \\ 
 
 \hline
 
 \textit{Manipulation (2 box, 2 target)*} & Move one of two boxes to one of 2 target locations.  & Ball & 2 & Proprio + Boxes + Targets + Task encoding & Proprio + Boxes \\ 
 
 \hline
 
 \textit{Manipulation (2 box gather)*} & Move two boxes close together.  & Ball & 2 & Proprio + Boxes + Targets + Task encoding & Proprio + Boxes \\ 
 \hline
 
\end{tabular}
\caption{\textbf{List of Tasks} A short summary of tasks used along with the information asymmetry used to train the \bpdot. Tasks marked with an asterisk are considered in Section \ref{sec:hier_experiments}. A more complete description is presented in Appendix \ref{appendix:tasks}.}
\label{table:task_list}
\end{table}

\begin{figure}%
    \centering
    \subfloat[All bodies considered. \label{fig:bodies}]{{\includegraphics[height=2.5cm]{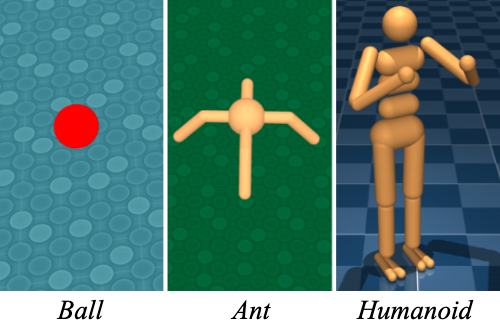} }}%
    \qquad
    \subfloat[Some of the tasks considered \label{fig:tasks}]{{\includegraphics[height=2.5cm]{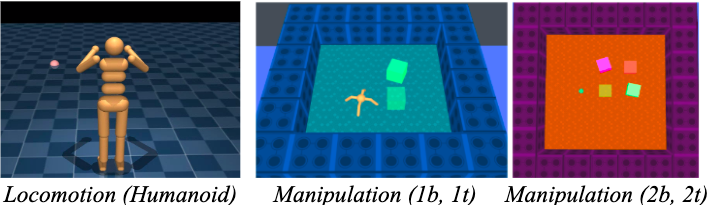} }}%
    \caption{\textbf{Visualization of tasks and bodies.} An illustration of some of the tasks and the bodies used for our experiments.}
    \label{fig:task_bodies}
\end{figure}

\subsection{Effect of Information Asymmetry}
\label{sec:info_asym_base}
The goal of this section is to both qualitatively and quantitatively understand the effect of modeling \textit{priors} with \emph{information constraints}.

To begin with, we focus our discussion on the \textit{Locomotion and Manipulation} task which is also considered in the hierarchical setting in Section \ref{sec:info_asym_2}. This task involves many behaviors like goal-directed movement and object interactions and thus provides a rich setting in which to study the effects of information asymmetry in some detail. We expand our discussion to a range of other tasks in Section \ref{sec:info_asym_rest}.

\subsubsection{Locomotion and Manipulation Task}
\label{sec:info_asym_1}
 This task involves a walker, a box and two goal locations. In order to solve the task, the agent must move the box to one of the goals (manipulation) and subsequently move to the other goal (locomotion). The locations of the agent, box and goals are randomized at the start of each episode. The agent receives a reward of +10 for successfully accomplishing either sub-task and an additional bonus  of +50 when it solves both. 
For this task we use an 8-DoF ant walker as shown in Figure \ref{fig:task_bodies}a.

The agent is given proprioceptive information, the location of the goals and box and the identity of the goal it must go to. The proprioceptive information includes the relative joint positions and velocities of the agent's body (details in Appendix \ref{appendix:bodies}). There are many choices for the information set sent to the \textit{prior} which affect the kinds of behaviors it can learn to model. For instance the \textit{prior} cannot capture the behavior of interacting with the box if it has no access to box information. On the other hand, a prior with access to all of the same information as the policy will learn to mimic the policy and lose the regularizing effect created by the information asymmetry. We examine whether this impacts performance on this task.

\begin{figure}%
    \centering
    \includegraphics[width=\linewidth]{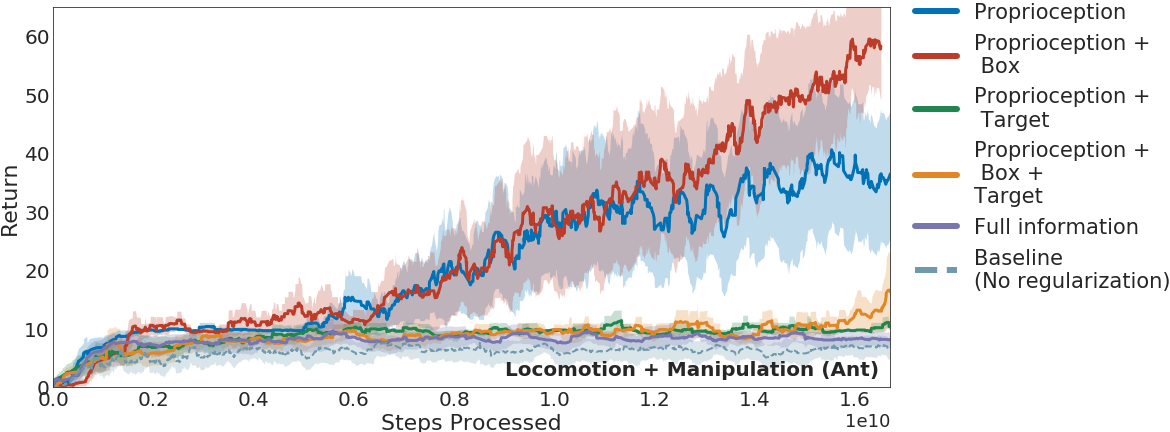}
    \caption{\textbf{Effect of information asymmetry.} Effect of learning with different levels of information asymmetry on the \textit{Locomotion and Manipulation} task. Label shows the kinds of information made accessible to the \textit{prior}.}
    \label{fig:info_asym_mbgtt}
\end{figure}

We present our results in Figure \ref{fig:info_asym_mbgtt}. We find a marked difference in performance based on the information that the \bp has access to. We observe that most policies on this task can easily get stuck in a local optima where the agent learns to solve the easier locomotion component of the task but never learns about the reward for moving the box. However, \bps can help regularize against this sub-optimal solution by encouraging useful patterns of behavior. For instance, a prior that has access to \textit{only} proprioceptive information learns to encode the space of trajectories containing primitive gaits or movements.
By encouraging the policy to continue to explore in this space, the agent is more likely to continue learning and to see the full task reward. In this case we find that the \bp that has access to both proprioception as well as the location of the box learns the fastest. 

\begin{figure}%
    \centering
    \includegraphics[width=\linewidth]{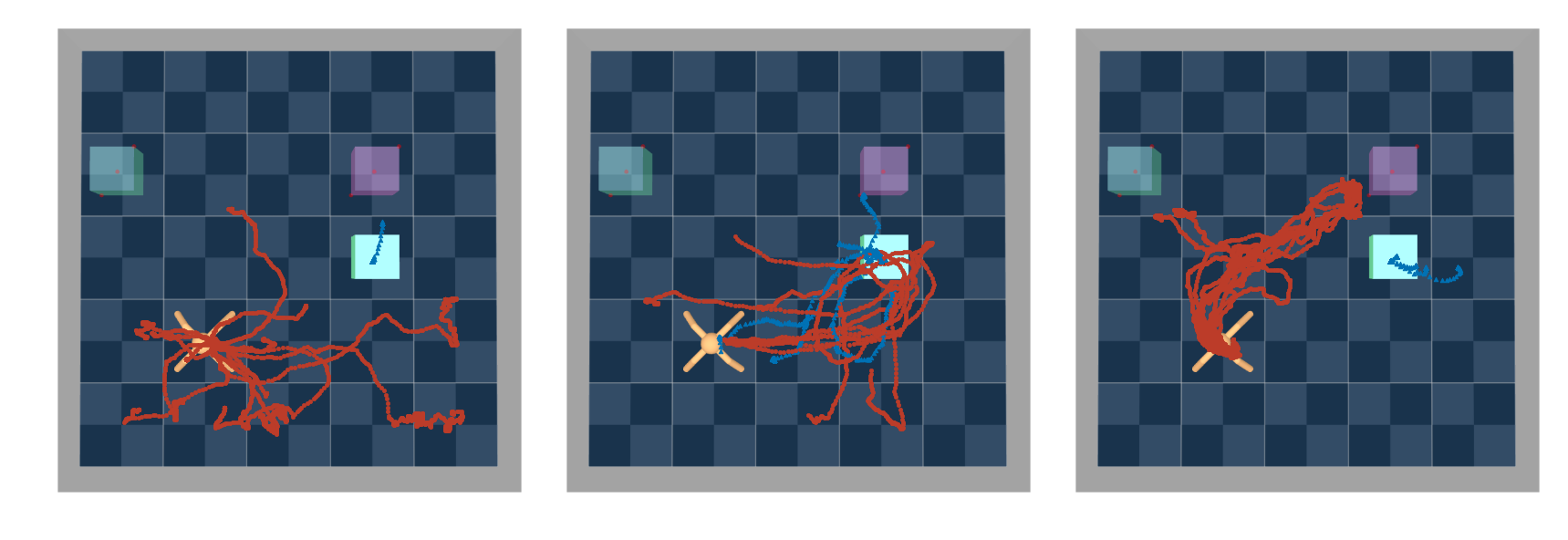}
    \caption{\textbf{Qualitative effect of information asymmetry.} Trajectories generated under \bps with access to different information sets for a fixed configuration of the box (solid blue), targets (semi-transparent blue and pink) and walker. The \bp has access to (left) proprioceptive information; (middle) proprioceptive information and the location of the box; (right) proprioceptive information and location of target. Red dotted line represents the movement of the ant and blue dotted line represents the movement of the box across multiple episodes.}
    \label{fig:ant_exploration}
\end{figure}

To understand why this might be, we perform a qualitative analysis by generating trajectories from some of the priors trained in Figure \ref{fig:info_asym_mbgtt}. Specifically, in Figure \ref{fig:ant_exploration}, we compare the kinds of behaviors that emerge when the prior has (left) only proprioceptive information; (middle) proprioception + box information and (right) proprioception + target locations. 
We observe that the prior with only proprioception learns to model a general space of movement where the agent explores the environment in an undirected manner. This behavior is already quite complex in that it involves specific patterns of joint actuation to move the body that can be modeled solely through information present in the proprioceptive state. In contrast to this, the prior with access to the box information learns to model behaviors related to moving the box around. The blue dotted line represents trajectories showing movement of the box as the agent interacts with it. Finally the agent with access to proprioception and the location of the targets (and not the box) also learns to model `goal-oriented' patterns of moving towards the different targets. 

It is worth pausing to understand this in more detail. While each instance of the task might involve a specific solution, like moving from point A to point B, together they all share a common behavior of `movement'. In fact these kinds of behaviors can be seen to occur at various levels of generality. For example, moving towards a specific object or target is less general than a goal-agnostic meander. The latter is more general and may thus be useful in new scenarios like maze exploration where they may not be any objects or targets. The analysis of Figure \ref{fig:info_asym_mbgtt} demonstrates that information constraints allow the \emph{prior} to generalize the behaviors it models. Moreover, different choices for the information set affect the kinds of behaviors modeled.

For this task it appears that the prior shown in Figure \ref{fig:ant_exploration}b performs the best but in general which of these behaviors is better depends on the setting we would like to evaluate them in. For instance, if we were to transfer to a task where interacting with the box results in a negative reward or where the goal is to move to one of the targets, then the prior in Figure \ref{fig:ant_exploration}c might work better than the one in Figure \ref{fig:ant_exploration}b. Alternatively a task where the optimal solution favors random exploration away from the box \emph{and} the targets would suit the prior in Figure \ref{fig:ant_exploration}a. The point here is to demonstrate that altering the information set that the prior has access to can lead to different \emph{kinds} of behaviors that are modeled under it and the best choice may depend on the domain we are ultimately interested in solving.

\subsubsection{Other Tasks}
\label{sec:info_asym_rest}

In this section, we expand our analysis of the effect of information asymmetry for learning to three more tasks: \textit{Manipulation (1 box, 3 targets)}, \textit{Locomotion (Humanoid)} and \textit{Locomotion (Ant)} (refer to Table \ref{table:task_list} for a short summary).

\begin{figure}%
    \centering
    \includegraphics[width=\linewidth]{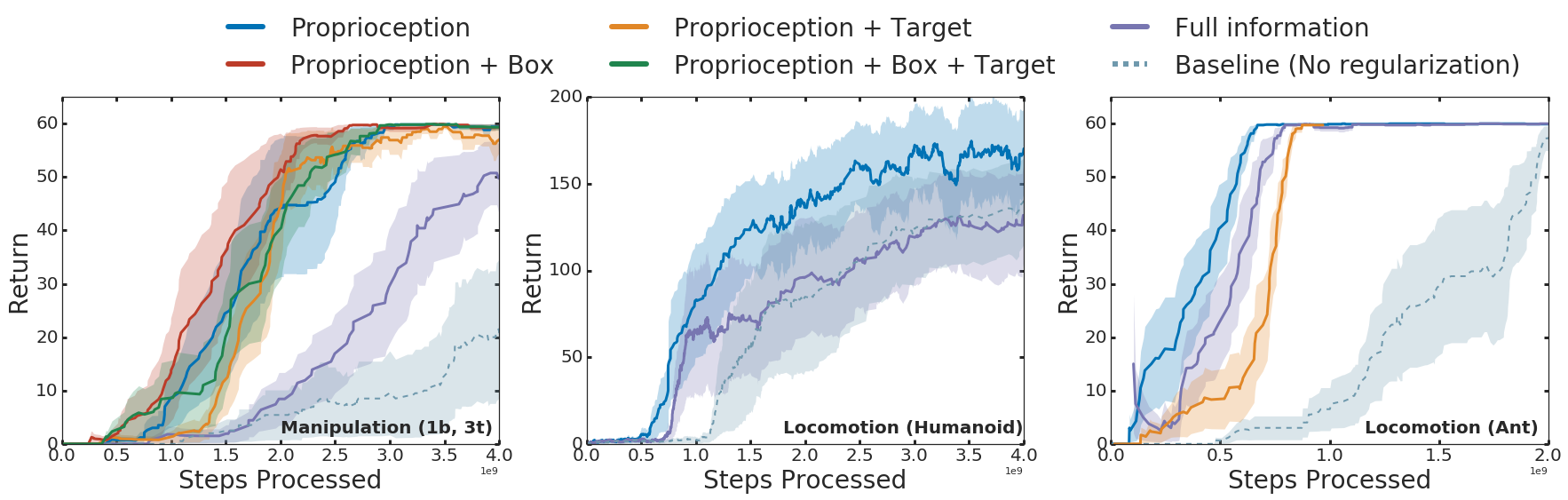}
    \caption{\textbf{Effect of information asymmetry.} Effect of learning with different levels of information asymmetry on \textit{Manipulation (1 box, 3 targets)}, \textit{Locomotion (Humanoid)} and \textit{Locomotion (Ant)}.}
    \label{fig:info_asym_rest}
\end{figure}

We show our findings in Figure \ref{fig:info_asym_rest}. We observe that regularizing against a \bp always provides a benefit and \textit{priors} with partial information  perform better than ones with full information. Additionally, \bps with access to only proprioceptive information tend to perform the best in tasks involving just locomotion.

\begin{figure}%
    \centering
    \subfloat[Trajectory of humanoid visiting targets in the \textit{Locomotion (Humanoid)} task under the optimal policy. \label{fig:gtt_humanoid_targets}]{{\includegraphics[scale=0.14]{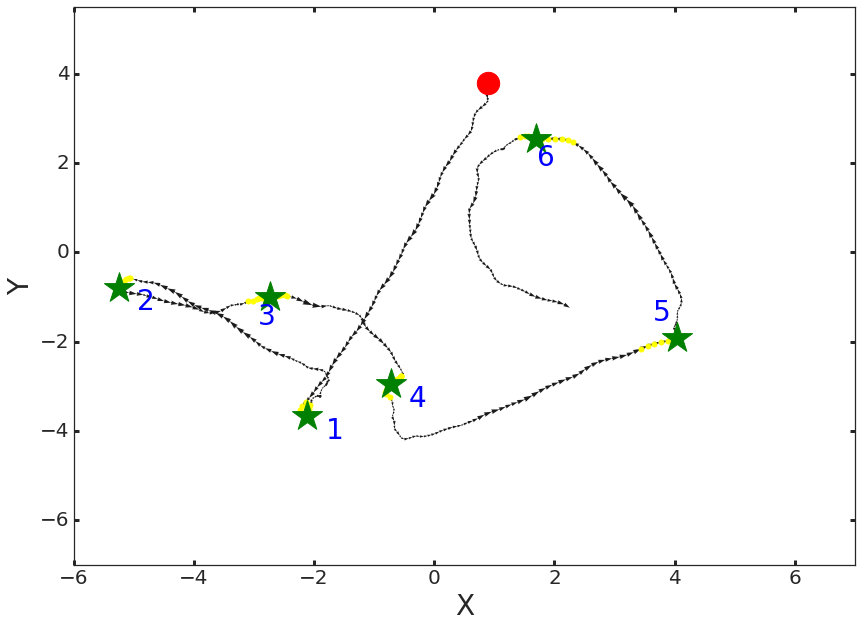} }}%
    \qquad
    \subfloat[Deviation of policy from the \bp in terms of KL divergence. \label{fig:gtt_humanoid_kl}]{{\includegraphics[scale=0.14]{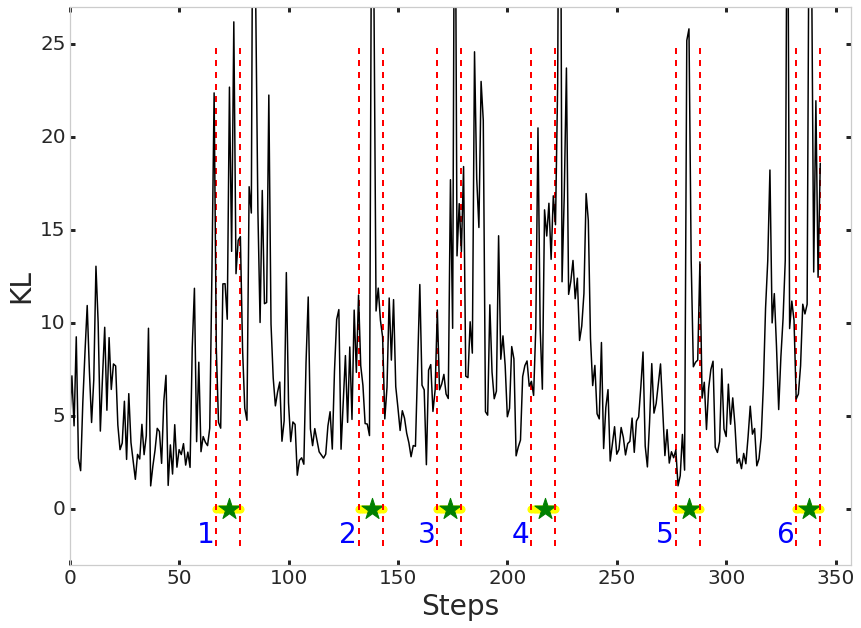} }}%
    \qquad
    \subfloat[Trajectories generated by the \bpdot. \label{fig:gtt_humanoid_prior}]{{\includegraphics[scale=0.14]{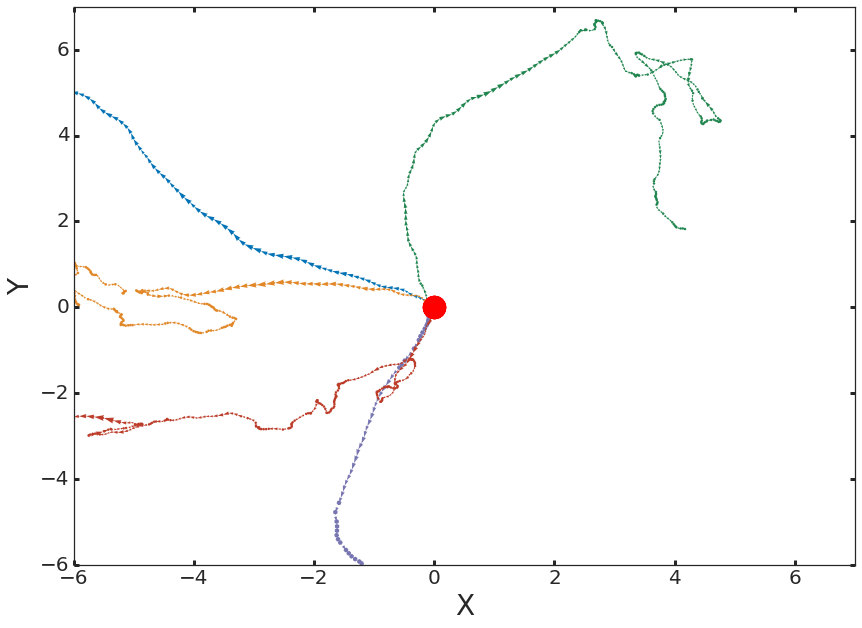} }}%
    \caption{\textbf{Analysis of the \textit{Locomotion (Humanoid)} task}. The spikes in KL divergence in the center align with the change in target as on the left; this is consistent with the idea that the \bp encodes primitive movement patterns. (right)  Trajectories generated under the \bp result in forward movement with occasional turns.}
    \label{fig:gtt_humanoid}
\end{figure}

In order to understand this in greater detail, we qualitatively analyze the \textit{Locomotion (Humanoid)} task in Figure \ref{fig:gtt_humanoid}. Our analysis demonstrates that the \emph{prior} with only proprioceptive information models forward movement in different directions with occasional sharp turns. In Figure \ref{fig:gtt_humanoid_kl} we plot the KL divergence between the \textit{prior} and policy as the agent visits different goals. Spikes in the KL divergence align precisely with the points \emph{where new targets} are chosen as shown in the trajectory in Figure \ref{fig:gtt_humanoid_targets}. 
As the figure illustrates, the \bp and the policy match quite closely (and so the KL is low) until the agent reaches a target. At this point, the policy changes to a new pattern of movement in order to change direction towards the new target location. As the humanoid settles into a new movement pattern, we observe that the KL once again reduces. This sequence repeats for each of the targets. Furthermore, in Figure \ref{fig:gtt_humanoid_prior}, we generate trajectories sampled from the learnt \textit{prior} starting from the same initial position. The \bp captures forward movement in different directions with occasional sharp turns. 

These findings are quite similar with our observations from the previous section. The \emph{prior} does not have access to the goal location. Therefore, it learns to mimic the experts behavior by capturing higher order statistics relating to movement that largely depends on the current configuration of the body. While this behavior is not perfect and mostly involves forward movements with occasional sharp turns, it demonstrates that simple MLP models can learn to capture general behaviors even for complex bodies like the 21-DoF humanoid robot.

\subsection{Sequential Navigation}
\label{sec:toy_domain_1}
\begin{figure}[t]
    \centering
    \subfloat[Exploration with a randomly initialized Gaussian policy. \label{fig:visualize_random}]{{\includegraphics[width=0.3\linewidth]{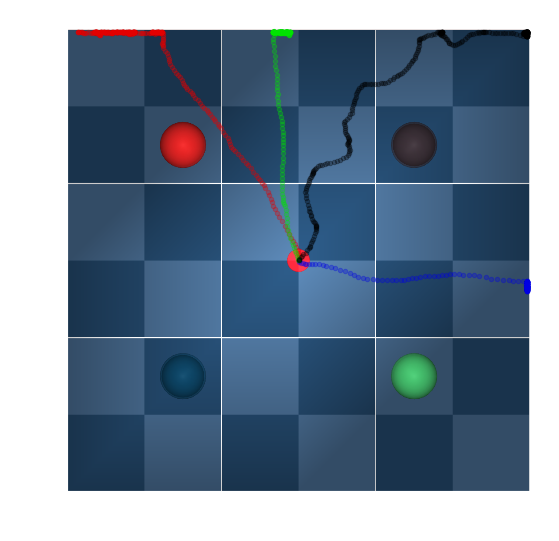} }}%
    \qquad
    \subfloat[Exploration with the learnt \bp model (LSTM). \label{fig:visualize_prior}]{{\includegraphics[width=0.3\linewidth]{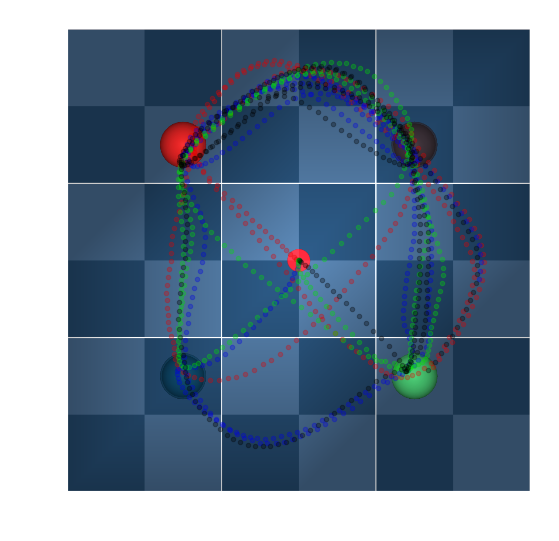} }}%
    \caption{\textbf{Visualization of exploration using a \bp}. Each colored dotted line represents a trajectory generated by rolling out a policy in the task.}
    \label{fig:visualize_traj}
\end{figure}

In the previous section we demonstrated that information constraints in the \textit{prior} affects the kinds of behaviors it learns and how this in turn can affect the speed of learning. In this section, we instead analyze how 
\emph{model capacity} (i.e. model architecture)
can affect the kinds of behaviors that are captured by a \bpdot. 
To this end, we study `Sequential Navigation' tasks that are designed to exhibit structure at multiple temporal and spatial scales and thus allow us to study behaviors at a coarser timescale. In these tasks an agent must learn to visit targets in a history-dependent sequence as illustrated by Figure \ref{fig:visualize_traj}. 

Specifically, sequences of targets are chosen based on 2nd order Markovian dynamics where some targets are more likely to be chosen given the two preceding targets than others (see Appendix \ref{appendix:environments} for details). In this setting the behavioral structure that could usefully be modeled by the \emph{prior} includes, effectively, a hierarchy of behaviors of increasing specificity: undirected locomotion behavior, goal-directed locomotion, as well as the long-horizon behavior that arises from some target sequences being more likely than others. 

We consider two kinds of models for our \bps - one that only uses a Multilayer Perceptron (MLP) architecture and another that incorporates an LSTM \citep[Long short-term memory:][]{lstm} and can thus learn to model temporal correlations. In both cases the policy and critic include LSTMs. Our goal is to analyze the kinds of behaviors each of these models can learn and understand the effect this has on learning and transfer.

We consider two versions of this task:
\begin{itemize}
    \item \textbf{Sequential Navigation Easy}: On each episode, a random sequence of targets is chosen based on the underlying Markovian dynamics. In this setting the agent is given the immediate next target to visit as input at each point in time.
    \item \textbf{Sequential Navigation Hard}: In this setting, the agent must learn to visit a single 4 target sequence (target 1 followed by target 2, 3 and then 4) and does not receive any additional information. Crucially, the sequence of targets used during transfer is \emph{more likely} to be generated under the Markovian dynamics used for training (compared to a uniform distribution over 4 target sequences). The agent receives a bonus reward of +50 for completing the task. In order to ensure that the task is fully observable under the MLP prior, we provide it with additional information indicating how far it has solved the sequence so far. The LSTM prior does not receive this additional information.
\end{itemize}

In both variants of the task, the agent receives the coordinates of all of the targets as well as its own location and velocity information. The \bps also receive this information but do not get the task specific information described above. For this section, we consider a version of this task with the 2-DoF pointmass body. The target locations are randomized at the start of each episode and the agent is given a reward of +10 each time it visits the correct target in the sequence. For the transfer experiment, since there is no additional information beyond what the \emph{prior} receives, we use the \emph{prior} to initialize the weights of the agent policy. Each training curve for transfer averages across 10 seeds (2 seeds for each of the 5 seeds from training). 

\begin{figure}%
    \centering
    \subfloat[Learning with priors on Easy Task\label{fig:easy_set_1}]{{\includegraphics[scale=0.28]{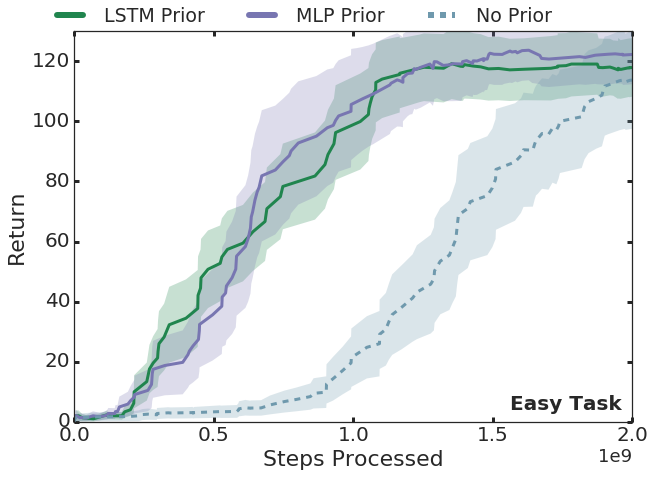} }}%
    \qquad
    \subfloat[Transfer of priors to hard
    task\label{fig:hard_set_1}]{{\includegraphics[scale=0.28]{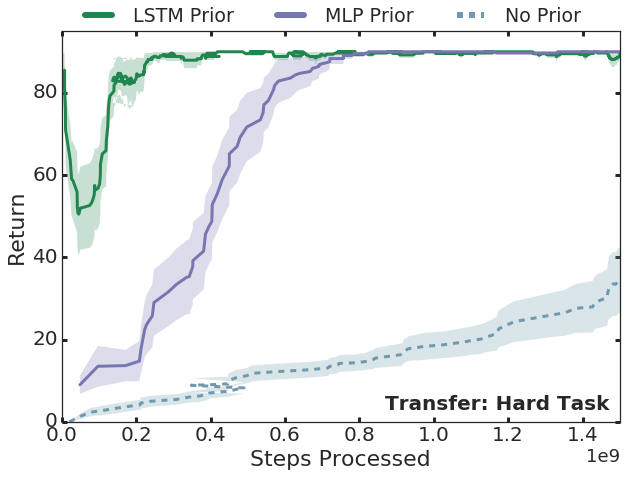} }}%
    \caption{\textbf{Learning and transfer on Sequential Navigation.} Learning curves from training on the easy task and transferring to the hard task with various \bp architectures.}
    \label{fig:pointmass_results}
\end{figure}

We illustrate our results in Figure \ref{fig:pointmass_results}. Figure \ref{fig:easy_set_1} shows that learning with either \textit{prior} leads to faster convergence on this task. This is in line with our intuition that the structure captured by the \textit{prior} can guide the learning process. Results on the transfer domain are shown in Figure \ref{fig:hard_set_1}. Our results show that learning is considerably accelerated when regularizing against either pretrained priors. Additionally, there is a significant improvement in learning speed when using the LSTM model. Since we do not transfer the critic and the SVG-0 algorithm relies on an informed critic to guide policy learning, we see a temporary drop in performance across all curves early in learning.

From these results we can conclude that the constraints defined by the model used for the \emph{prior} clearly affect the kinds of behaviors learnt by it. A model that can capture the temporal structure present in the training task is \textit{more likely} to generate the target sequence for transfer. We analyze the general behaviors captured under each \textit{prior} below.

\subsubsection{Analysis}
To understand why the \bps are useful during training and transfer we first need to clarify what \emph{behaviors} they are actually modelling. In order to do this, we qualitatively compare trajectories generated from the trained MLP prior to `random' trajectories generated from a randomly initialized Gaussian policy as shown in Figure \ref{fig:visualize_traj}.

As the figure shows, the \textit{prior} learns to model a `goal-directed' behavior of bouncing between the different targets as opposed to the random motion of the untrained policy. Since the prior does not have access to the task-specific information of \textit{which target} to visit next, it learns to encode the behavior of visiting targets in general. This `target-centric' movement behavior is useful to guide the final policy during training and transfer. 

\begin{figure}%
    \centering
    \includegraphics[scale=0.26]{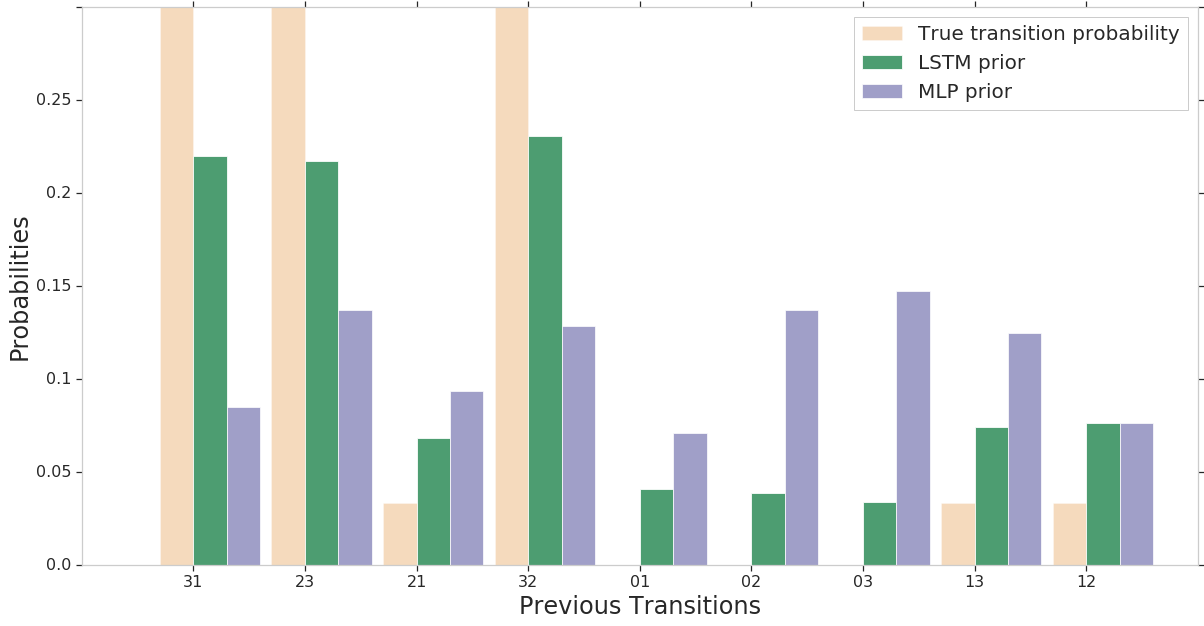}
    \caption{\textbf{Transition dynamics} generated under different \bp models to visit target 0. Each point on the x-axis indicates the indices of the last two targets visited. On the y-axis we plot the probability of visiting target 0 for each transition sequence.}
    \label{fig:transition_dynamics}
\end{figure}

The LSTM prior exhibits goal directed behavior similar to the MLP prior but additionally models the long-range structure of the domain. To understand 
this difference, we can compare the transition dynamics when they visit a specific target. In Figure \ref{fig:transition_dynamics}, we plot the empirical distribution of visiting target 0 on the Y-axis against the last 2 visited targets on the X for each model. The distribution generated under the LSTM closely resembles the true underlying distribution used to generate the target sequence. The MLP prior instead learns to visit each target more or less uniformly at random. This results in a significant difference in performance during transfer where the LSTM model is more likely to generate a trajectory that visits targets in the rewarded sequence.

It is important to note that restricting the space of trajectories that the prior explores is only advantageous if that space also captures the \textit{solution space} of the task we are ultimately interested in solving. In other words, there is a tradeoff between generality and efficacy of the prior. A prior over trajectories that is `uniform' in that it assigns equal mass to all possible trajectories may be of limited use since it does not capture any of the underlying task structure. On the other hand, a prior that overfits to the dynamics of a specific task will not be general enough to be useful on many tasks. Understanding how the choice of model and information set affect this tradeoff is thus crucial to learn effective priors.

\section{Structured \bp models}
\label{sec:structured_models}

In the discussion thus far, we have considered examples where $\pp$ and $\qp$ use neural network models to parameterize uni-modal Gaussian distributions; a form which has shown promising results but is relatively simple in its \emph{stochastic structure} and in the kinds of trajectory distributions it can model. 

To see why such a simple model might be limiting, consider an example where the optimal action in a particular state depends on some unobserved context; for instance an agent may have to go left or  right based on whether a light is on. If we were to model this on a continuous spectrum going `left' might correspond to an action value of -1 and going `right' to a value of +1. A \emph{prior} without  access to the conditioning information (the light in our example) would need to model the marginal distribution, which, in this case, is bi-modal (with one mode at -1 and the other at +1). It is impossible to capture such a distribution with a uni-modal Gaussian. Due to nature of the objective (cf.\ Equation\ \ref{eq:ProbabilisticRL:OptimalPriorRaw}) the \emph{prior} will have to cover both modes. This might lead to undesirable solutions where the mean (and mode) of the \textit{prior} is centered on the `average' action of 0.0, which is never a good choice. 

While this example may seem concocted, it does illustrate that there may be situations where it will be important to be able to model more complex distributions. In fact, as discussed in Section \ref{sec:BP}, the optimal \emph{prior} in the multi-task setting is a mixture of the solutions to individual task (cf.\ Equation \ref{eq:ProbabilisticRL:OptimalPrior}). This may not be modelled well by a Gaussian.

In this section we will 
develop the mathematical tools required to 
work with more complex models. Our starting point, is to once again turn to the probabilistic modeling literature to borrow ideas from the space of latent variable models. 
Latent or `unobserved' variables in a probabilistic model can be used to increase capacity, introduce inductive biases and model complex distributions. In the most general form we can consider directed latent variable models for both $\pp$ and $\qp$ of the following form:
\begin{align}
\pp(\tau) &= \textstyle \int \pp(\tau| y) \pp(y) d y, \label{eq:latent_formulation_prior}\\
\qq(\tau) &= \textstyle \int \qq(\tau | z) \qq(z) d z
\label{eq:latent_formulation}
\end{align}
where the unobserved `latents' $y$ and $z$ can be time varying, e.g.\ $y=(y_1, \dots y_T)$, continuous or discrete, and can exhibit further structure. 

Above, we have motivated latent variables in the \emph{prior} $\pp$. To motivate \emph{policies} $\qp$ with latent variables, we can consider tasks that admit multiple solutions that achieve the same reward. The $\KL$ term towards a suitable prior can create pressure to learn a distribution over solutions (instead of just a single trajectory), and augmenting $\qp$ may make it easier to model these distinct solutions \citep[e.g.][]{hausman2018learning}. 

Policies with latent variables have also been considered within the RL community under the umbrella of hierarchical reinforcement learning (HRL). While the details vary, the motivation is often to model  higher level abstract actions or `options' ($z_t$s in our formulation). These are often be temporally consistent, i.e. the abstract actions may induce correlations between the primitive actions. This may be beneficial, for instance for exploration and credit assignment on long horizon tasks. The focus of the present work is for modeling \bps $\pp$, i.e.\ we are interested in models that can provide \emph{richer descriptions of behavior}. We discuss the relations between our method and some work from HRL in Section \ref{sec:related}. 

Unfortunately, a direct use of the formulation from Equations (\ref{eq:latent_formulation_prior}) and (\ref{eq:latent_formulation}) can be problematic. It is often difficult to compute the KL term in Equation (\ref{eq:objective:generic_pertimestep}): $\KL[ \qp(\tau) || \pp(\tau) ]$ when $\qp$ and $\pp$ are \emph{marginal} distributions of the form outlined in Equation (\ref{eq:latent_formulation}) and Equation (\ref{eq:latent_formulation_prior}) respectively. This results in the need for approximations and a number of choices and practical decisions stem from this consideration. We focus the discussion here on a simple and practical formulation that is the focus of our experimental analysis in Section \ref{sec:hier_experiments}. A more detailed discussion of the effects of various modeling choices is deferred to Section \ref{sec:related}. 

\subsection{Simplified Form}
In this work we focus on a formulation that allows for continuous latent variables in \emph{both} $\pp$ and $\qp$ which is both tractable and practical to implement. The formulation allows $\qp$ and $\pp$ to model richer (state-conditional) action distributions, to model temporal correlations, and provides flexibility for partial parameter sharing between $\pp$ and $\qq$.

We consider a continuous latent variable $z_t$ which has the same dimension and semantics in $\qp$ and $\pp$. We divide $\qp$ into higher level (HL) $\qp^H(z_t| x_t)$ and lower level (LL)  $\qp^L(a_t| z_t, x_t)$ components. We can then derive the following bound for the KL (see Appendix \ref{appendix:hierarchical_objective_derivation} for proof):
\begin{align}
\KL[\qp(a_t|x_t) || \pp(a_t|x_t)]
&\le \KL[\qp^H(z_t|x_t) || \pp^H(z_t|x_t)] \nonumber \\
&\hspace{1.8cm}\textstyle + \EE_{\qp^H(z_t|x_t)}[\KL[\qp^L(a_t|z_t, x_t) || \pp^L(a_t|z_t, x_t)] ]
\label{eq:objective:hierarchical}
\end{align}
which can be approximated via Monte Carlo sampling and we now define $x_t$ such as to contain previous $z_t$s as well. This upper bound effectively splits the $\KL$ into two terms - one between the higher levels $\qp^H(z_t| x_t)$ and $\pp^H(z_t| x_t)$ and the other between the lower levels $\qp^L(a_t| x_t, z_t)$ and $\pp^L(a_t| x_t, z_t)$. 

\subsection{Modeling considerations}
In the remainder of this section, we describe modeling considerations that are needed in order to implement the KL-regularized objective using Equation (\ref{eq:objective:hierarchical}).

\paragraph{Regularizing using information asymmetry} 
\label{sec:hierarchy:info_asym}
In Section \ref{sec:info_asym_1}, we demonstrated that information constraints 
can force the prior to generalize across tasks or different parts of the state space.
This intuition 
also applies
to the different levels of a hierarchical policy. The constraint between the higher level policies in Equation (\ref{eq:objective:hierarchical}) has two effects: it regularizes the higher level action space making it easier to sample from; and it introduces an information bottleneck between the two levels. The higher level thus pays a `price' for every bit it communicates to the lower level. This encourages the lower level to operate as independently as possible to solve the task. By introducing an information constraint on the lower level, we can force it to model a general set of skills that are modulated via the higher level action $z$ in order to solve the task. Therefore, the tradeoff between generality and specificity of \bps mentioned in Section \ref{sec:experiments} would now apply to the different layers of the hierarchy. We will explore the effect of this empirically in Section \ref{sec:hierarchy:info_asym}.

\begin{figure}%
    \centering
    \includegraphics[scale=0.3]{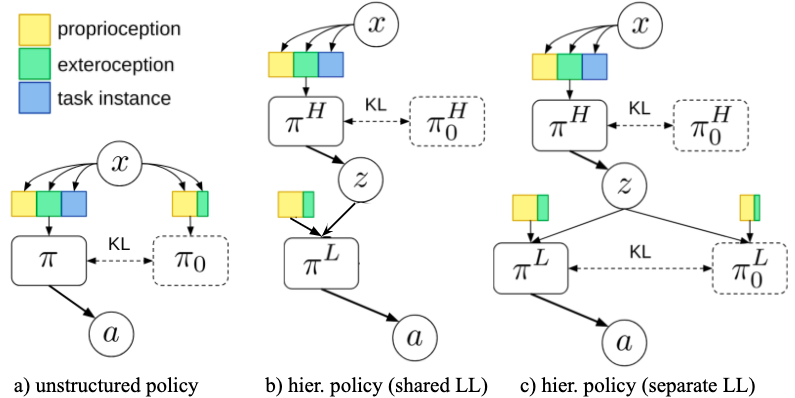}
    \caption{\textbf{Training setup for structured and unstructured models.}}
    \label{fig:structured_priors_descr}
\end{figure}

\paragraph{Partial parameter sharing}
\label{sec:hierarchy:sharing}
An advantage of the hierarchical structure is that it enables several options for partial parameter sharing, which when used in the appropriate context can make learning more efficient. For instance, sharing the lower level controllers between the agent and default policy, i.e.\ setting $\qp^L(a_t|z_t,x_t)=\pp^L(a_t|z_t,x_t)$ reduces the number of parameters to be trained and allows skills to be directly reused. This amounts to a hard constraint that forces the KL between the lower levels to zero. With that, the objective from Equation (\ref{eq:objective:info_asym}) now becomes:
\begin{align}
\mathcal{L}(\qp, \pp) & \textstyle \geq \textstyle \EE_{\tau}\left[
\sum_{t\ge1} \gamma^t r(s_t,a_t)
    - \alpha\gamma^t\KL[\qp(z_t | x_t) || \pp(z_t | x_t)] \right]
    \label{eq:latent_lower_bound}
\end{align}
where the KL divergence is between policies defined only on abstract actions $z_t$. We illustrate the various approaches to training structured and unstructured models with shared and separate lower levels in Figure \ref{fig:structured_priors_descr}.

\subsection{Algorithmic considerations}
Off-policy training of hierarchical policies adds complexity in the estimation of the state-action value $Q(x_t, a_t, z_t)$, the bootstrap value $V(x_t, z_t)$ and for learning the policy itself. We present a modified version of Algorithm \ref{alg:rs0_distral} for hierarchical policies in Algorithm \ref{alg:r2h2}, where the modifications required for learning \bps are in blue and those for training latent variable models are in purple. The algorithm assumes the latent $z$ is sampled at every timestep. There are additional complexities when the latent is held fixed across multiple timesteps \citep[e.g. similar to][]{heess2016learning}, which we describe in greater detail below.

\paragraph{Critic update:} We adapt the retrace algorithm from \citet{munos2016safe} for off-policy correction to hierarchical policies. At timestep $t$, the update is given by:
\begin{align}
  Q(x_{t}, a_{t}, z_t) &= Q(x_{t}, a_{t}, z_t)
        +
        \sum_{s \ge t}\gamma^{s-t}\left(\prod_{i=t}^{s}c_i\right)
        \left(r_s+\gamma V_{s+1} - Q(x_s, a_s, z_s)\right) \nonumber \\
 c_{i}&=\lambda \min\left(\frac{\pi^H(z_{i}| x_{i}) \pi^L(a_{i}| x_i, z_i)}{\mu(a_{i}| x_{i})}, 1\right) \nonumber
\end{align}   
Note that the state-action value function $Q$ also depends on the latent $z_t$ here. This is because in the general case, $z_t$ may be sampled infrequently or held constant across multiple timesteps. In other words, the value of \emph{future} states $x_s$ where $s > t$ could depend on the current latent $z_t$ and must be accounted for.

This may also hold for the state-value function based on whether $z_t$ is sampled in state $x_t$. Specifically, in states where $z_t$ is sampled:
\begin{align}
  V(x_{t}) &= \EE_{z\sim \qp^H(.| x_t), a \sim \qp^L(.|x_t, z)}Q(x_{t}, a, z) \nonumber
\end{align}   
and in other states:
\begin{align}
  V(x_{t}, z_t) &= \EE_{a \sim \qp^L(.|x_t, z_t)}Q(x_{t}, a, z_t) \nonumber
\end{align}   

Additionally, in our adaptation of the retrace estimator, we do not consider the latent $z_t$ in the behavior policy $\mu$ that was used to generate the data on the actors. Since the latent does not affect the environment directly, we can estimate the importance weight as $\mu(a_t | x_t)$ (as a way to reduce the variance of the estimator). 

\paragraph{Policy update:} In the hierarchical formulation there are different ways to adapt the policy update from the SVG-0 algorithm \citep{heess2015learning}. 

One way to compute the update for $\qp^H$ is through the partial derivative of $Q$:
\begin{align}
\nabla_{\theta^H} \EE_{\qp} \left[ Q(x, a, z) \right] &= \EE_{\zeta(\eta)} \left [\frac{\partial Q}{\partial z} \nabla_{\theta^H} \qp (a| x, \eta) \right] \nonumber
\end{align}
where $z \sim \pi^H(z|x)$ is reparameterized as $z = \pi^H(z| x, \eta)$, for some noise distribution $\eta \sim \zeta(.)$. In practice we found that this leads to worse performance and opted instead to compute the policy update as:
\begin{align}
\nabla_{\theta} \EE_{\qp} \left[ Q(x, a, z) \right] &= \EE_{\zeta(\eta), \rho(\epsilon)} \left [\frac{\partial Q}{\partial a} \nabla_{\theta^H} \qp (z| x, \eta) \nabla_{\theta^L} \qp (a| s, z, \epsilon) \right] \nonumber
\end{align}
where we introduce two sources of noise $\eta \sim \zeta(.)$ and $\epsilon \sim \rho(.)$. Finally the policy update also includes the KL terms from the objective as in Algorithm \ref{alg:rs0_distral} using the bound from Equation (\ref{eq:latent_lower_bound}).

\paragraph{Prior update:} The update for the \bp is exactly as in Algorithm \ref{alg:rs0_distral} except that here we use the bound from Equation (\ref{eq:latent_lower_bound}) to compute the $\KL$ used for distillation. Intuitively, this can be thought of as two separate distillation steps - one for the HL and the other for the LL.

\begin{algorithm*}[tb]
  \caption{SVG(0) with experience replay for hierarchical policy}
  \label{alg:r2h2}
\begin{algorithmic}
  \scriptsize
  \State Flat policy: $\pi_\theta(a_t|\epsilon_t,x_t)$ with parameter $\theta$
  \State HL policy: {\color{purple} $\pi^H_\theta(z_t|x_t)$, where latent is sampled by reparameterization $z_t=f^H_\theta(x_t, \epsilon_t)$}
  \State \textit{Behavioral Priors}: {\color{blue} $\pi^H_{0,\phi}(z_t|x_t)$ and $\pi^L_{0,\phi}(a_t|z_t,x_t)$ with parameter $\phi$}
  \State Q-function: {\color{purple} $Q_\psi(a_t,z_t,x_t)$ with parameter $\psi$}
  \\
  \State Initialize target parameters $\theta^\prime \leftarrow \theta$,
  \hspace{0.2cm} {\color{blue} $\phi^\prime \leftarrow \phi$}, \hspace{0.2cm} $\psi^\prime \leftarrow \psi$.
  \State Hyperparameters: {\color{blue} $\alpha$}, $\alpha_H$, $\beta_{\qp}$, {\color{blue}$\beta_{\pp}$}, $\beta_{Q}$
  \State Target update counter: $c \leftarrow 0$
  \State Target update period: $P$
  \State Replay buffer: $\mathcal{B}$
  \\
  \For{$j=0, 1, 2, ... $}
  \State Sample partial trajectory $\tau_{t:t+K}$ generated by behavior policy $\mu$ from replay $B$:\\ \hspace{1.6cm}$\tau_{t:t+K}=(s_t,a_t,r_t,...,r_{t+K})$,\\
  \For{$t^\prime=t, ...t+K$}\\
  \hspace{1.6cm} {\color{purple} $\epsilon_{t^\prime} \sim \rho(\epsilon)$, $z_{t^\prime}=f^H_\theta(x_{t^\prime},\epsilon_{t^\prime})$}\\
  \Comment{Sample latent via reparameterization} 
  \\
  \hspace{1.6cm}{\color{blue} $\hat{\KL}_{t^\prime}=\KL\left[\pi^H_\theta(z|x_{t^\prime})\|\pi^H_{0,{\phi^\prime}}(z|x_{t^\prime})\right]+
  \KL\left[\pi^L_\theta(a|z_{t^\prime},x_{t^\prime})\|\pi^L_{0,\phi^\prime}(a|z_{t^\prime},x_{t^\prime})\right]$}
  \Comment{Compute KL}\\ 
  \hspace{1.6cm}{\color{blue} $\hat{\KL}^\mathcal{D}_{t^\prime}=\KL\left[\pi^H_\theta(z|x_{t^\prime})\|\pi^H_{0,{\phi}}(z|x_{t^\prime})\right]+
  \KL\left[\pi^L_\theta(a|z_{t^\prime},x_{t^\prime})\|\pi^L_{0,\phi}(a|z_{t^\prime},x_{t^\prime})\right]$}\\
  \Comment{Compute KL for Distillation}\\   
  \hspace{1.6cm}$\hat{\text{H}}_{t^\prime}=\EE_{\pi_\theta(a|\epsilon_{t^\prime},x_{t^\prime})}[\log\pi_\theta(a|\epsilon_{t^\prime},x_{t^\prime})]$  
  \Comment{Compute action entropy}\\
  \\
  \hspace{1.6cm} $\hat{V}_{t^\prime}
        =\EE_{\pi_\theta(a|\epsilon_{t^\prime},x_{t^\prime})}
        {\color{purple}\left[Q_{\psi^\prime}(a,z_{t^\prime},x_{t^\prime})\right]}$-${\color{blue}\alpha\hat{\KL}_{t^\prime}}$
  \Comment{Estimate bootstrap value}\\
  \hspace{1.6cm}$\hat{c}_{t^\prime}=\lambda \min\left(\frac{\pi_\theta(a_{t^\prime}|\epsilon_{t^\prime},x_{t^\prime})}{\mu(a_{t^\prime}| x_{t^\prime})}\right)$
  \Comment{Estimate traces~\citet{munos2016safe}}\\  
  \hspace{1.6cm}
  $\hat{Q}^R_{t^\prime}={\color{purple}Q_{\psi^\prime}(a_{t^\prime},z_{t^\prime},x_{t^\prime})}
        +
        \sum_{s \ge t^\prime}\gamma^{s-t^\prime}\left(\prod_{i=s}^{t^\prime}\hat{c}_i\right)
        \left(r_s+\gamma \hat{V}_{s+1} - {\color{purple}Q_{\psi^\prime}(a_s,z_s,x_s)}\right)$\\
  \Comment{Apply Retrace to estimate Q targets~\citet{munos2016safe}}\\
  \EndFor \\  
  \\
  \hspace{1.6cm}{\color{purple} $\hat{L}_Q=\sum_{i=t}^{t+K-1}\|\hat{Q}^R_i-Q_\psi(a,z_{i},x_{i})\|^2$}
  \Comment{Q-value loss}\\
  \\
  \hspace{1.6cm}$\hat{L}_\pi=\sum_{i=t}^{t+K-1}\EE_{\pi_\theta(a|\epsilon_{i},x_{i})}{\color{purple}Q_{\psi^\prime}(a,z_{i},x_{i})}-{\color{blue}\alpha\hat{\KL}_i} + \alpha_\text{H}\hat{\text{H}}_i$
  \Comment{Policy loss} \\
  \\
  \hspace{1.6cm}{\color{blue} $\hat{L}_{\pi^H_0}=\sum_{i=t}^{t+K-1}\hat{KL}_i^\mathcal{D}$}
  \Comment\textit{Prior} loss\\  
  \State
  $\theta \leftarrow \theta+\beta_\pi\nabla_\theta\hat{L}_\pi$ \hspace{0.3cm}
  {\color{blue} $\phi \leftarrow \phi+\beta_{\pi^H_0}\nabla_\phi\hat{L}_{\pi^H_0}$}
  \hspace{0.3cm}
  {\color{purple}$\psi \leftarrow \psi-\beta_{Q}\nabla_\psi\hat{L}_{Q}$}
  \State Increment counter $c \leftarrow c+1$
  \If{$c > P$}
  \State Update target parameters $\theta^\prime \leftarrow \theta$,
  \hspace{0.2cm} {\color{blue} $\phi^\prime \leftarrow \phi$}, \hspace{0.2cm} {\color{purple}$\psi^\prime \leftarrow \psi$}
  \State $c \leftarrow 0$
  \EndIf
  \EndFor
\end{algorithmic}
\end{algorithm*}
\section{Experiments with structured priors}
\label{sec:hier_experiments}

In this section, we analyze the performance of structured and unstructured \bps on a range of control domains. Our goal is to understand if hierarchy adds any advantages to learning compared to unstructured architectures, and to study the potential benefits of partial parameter sharing between $\qp$ and $\pp$ during transfer. We begin with an analysis similar to the one in Section \ref{sec:info_asym_1} to study how information asymmetry in modular policies affect the kinds of behaviors they learn to model.
Next we compare the performance of structured and unstructured models across a range of `Locomotion' and `Manipulation' tasks to analyze their benefit for training and transfer. Finally we revisit the `Sequential Navigation' task from Section \ref{sec:toy_domain_1} to study how architectural constraints affect the behaviors modeled by hierarchical policies.

Unless otherwise specified, all the structured priors we consider for this part include a shared lower level policy as described in Section \ref{sec:structured_models}. During transfer, this enables the reuse of skills captured within the lower level and restricts the parameters that are trained to those just in the higher level policy: $\qp^H$. 

We primarily consider two models for the higher level prior $\pp^H$:
\textbf{Independent isotropic Gaussian.} $\pi^H_0(z_t | x_t) = \mathcal{N}(z_t | 0, 1)$, where abstract actions are context independent or 
\textbf{AR(1) process.} $\pi^H_0(z_t | x_t) = \mathcal{N}(z_t | \alpha z_{t-1}, \sqrt{1-\alpha^2} )$, a first-order auto-regressive process with a fixed parameter $0 \leq \alpha < 1$ chosen to  ensure the variance of the noise is marginally distributed according to $\mathcal{N}(0, I)$. We include this model since it showed promising results in \cite{merel2018neural} and could allow for temporal dependence among the abstract actions. 
For the `Sequential Navigation' tasks considered in Section \ref{sec:toy_domain_2}, we consider two architectural variants of the prior with and without memory: one with an MLP (\textit{Hier.\ MLP}); and one with an LSTM (\textit{Hier.\ LSTM}). 

\subsection{Effect of Information Asymmetry}

\label{sec:info_asym_2}
As described in Section \ref{sec:hierarchy:info_asym}, we expect information asymmetry between the different levels of the hierarchy to play an important role in shaping the learning dynamics; an idea which we demonstrate empirically in this section.

Consider the \textit{Locomotion and Manipulation} task (see Table \ref{table:task_list}) from Section \ref{sec:info_asym_1} where an agent must move a box to a target (\textit{Manipulation}) and then move to another target (\textit{Locomotion}). This task has many options for information that can be withheld from the lower level allowing us to qualitatively and quantitatively analyze how this affects learning. For this experiment, we consider structured models with a shared lower level policy and an isotropic Gaussian higher level prior. 

\begin{figure}%
    \centering
    \includegraphics[width=\linewidth]{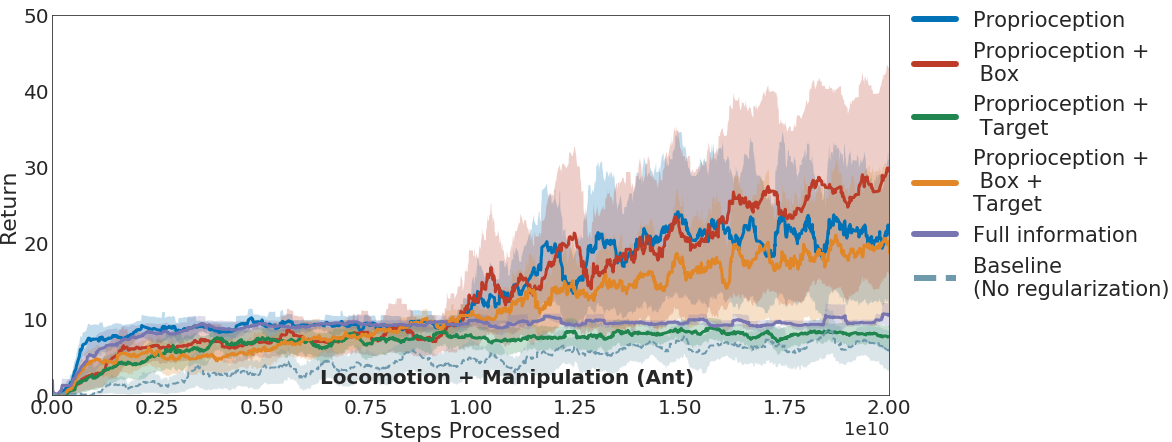}
    \caption{\textbf{Effect of information asymmetry} with structured models on the `Locomotion and Manipulation' task.}
    \label{fig:info_asym_mbgtt_hier}
\end{figure}

As Figure \ref{fig:info_asym_mbgtt_hier} illustrates, we see a similar effect of information asymmetry here as in Section \ref{sec:info_asym_1}. Specifically, \bps with partial information perform \emph{better} than ones with access to full information. As discussed previously, this task includes a local optimum where agents that learn to solve the easier `Locomotion' component of the task may not go on to solve the full task. \emph{Priors} with access to only partial information can help regularize against this behavior.

\begin{figure}%
    \centering
    \includegraphics[scale=0.25]{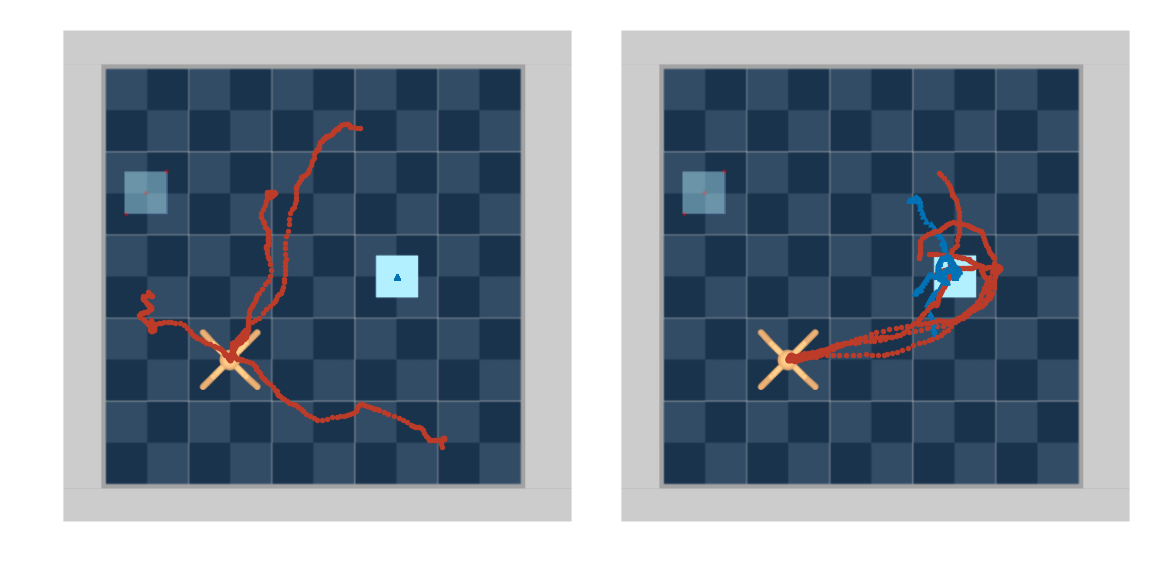}
    \caption{\textbf{Effect of information asymmetry on exploration} A visualization of the motion of the ant (red) and the box (blue) under random latent samples fed to a lower level policy with a) just proprioceptive information and b) proprioceptive and box information.}
    \label{fig:ant_rollout}
\end{figure}

To qualitatively understand these results we analyze trajectories generated using trained priors where the lower level has access to a) only proprioceptive information, and b) proprioception and box information. For each setting, we generate trajectories by sampling from the shared lower level policy conditioned on latents sampled from the higher level prior. 

We present our results in Figure \ref{fig:ant_rollout}.
Qualitatively, the trajectories are very similar to those observed in Section \ref{sec:info_asym_1}. The \textit{prior} with only proprioceptive information learns to model movement behaviors in different directions. In contrast, the \textit{prior} with access to the box location shows a more \textit{goal-directed} behavior where the agent moves toward the box and pushes it. 
However, compared to the unstructured priors of Section \ref{sec:experiments}, the hierarchical model structure used in the present section makes it easy to modulate the behavior using the latent variable $z$. 
The skills represented by the shared low-level can therefore more easily be specialized to a new task by training a new high-level policy to set $z$ appropriately.
We analyze this in the next section.

\subsection{Comparison of structured and unstructured priors}
\paragraph{Training:} We consider training performance on three tasks: \textit{Locomotion (Ant)}, \textit{Manipulation (1 box, 1 target)} and \textit{Manipulation (2 boxes, 2 targets)} (refer to Table \ref{table:task_list} for a short summary and the information asymmetry used). 

\begin{figure}%
    \centering
    \includegraphics[width=\linewidth]{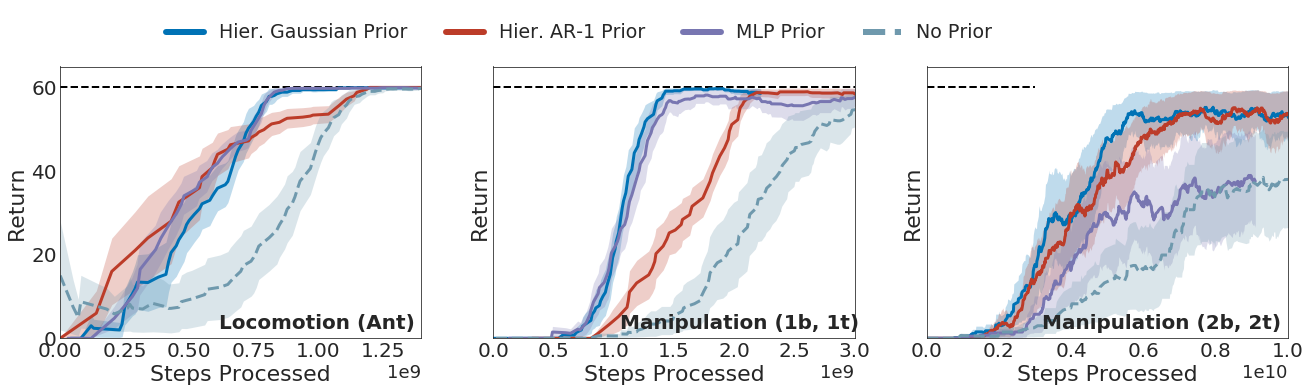}
    \caption{\textbf{Learning performance} with different \bps on the \textit{Locomotion (Ant)}, \textit{Manipulation (1b, 1t)} and \textit{Manipulation (2b, 2t)} tasks.}
    \label{fig:speedup_hier}
\end{figure}

We illustrate our results in Figure \ref{fig:speedup_hier}. They are consistent with Section \ref{sec:experiments}: \bps with partial information accelerate learning across all tasks. Furthermore, we find that hierarchical models tend to perform better than the unstructured ones on more complex tasks (\textit{Manipulation (2b, 2t)}). In contrast to the findings of \citet{merel2018neural}, we find that the AR-1 prior performs worse than the Gaussian prior. The AR-1 prior allows to model correlations across time. However, this added flexibility does not seem to offer an advantage when modeling skills required to solve these tasks.

A key motivation behind the introduction of the structured models was their ability to better capture complex distributions. In order to understand whether this effect might explain some of the performance differences we observed, we computed the average KL divergence between the policy and \textit{behavior prior} on the \textit{Manipulation (1 box, 1 target)} task across 5 seeds and 100 episodes. Values of 2.64 and 11.35 for the Hier.\ Gaussian and MLP prior are consistent with the idea that the latent variable model  provides a better \textit{prior} over the trajectory distribution and thus allows for a more favorable trade-off between KL and task reward. 

\begin{figure}%
    \centering
    \subfloat[ \label{fig:distrbn_distral}]{{\includegraphics[scale=0.13]{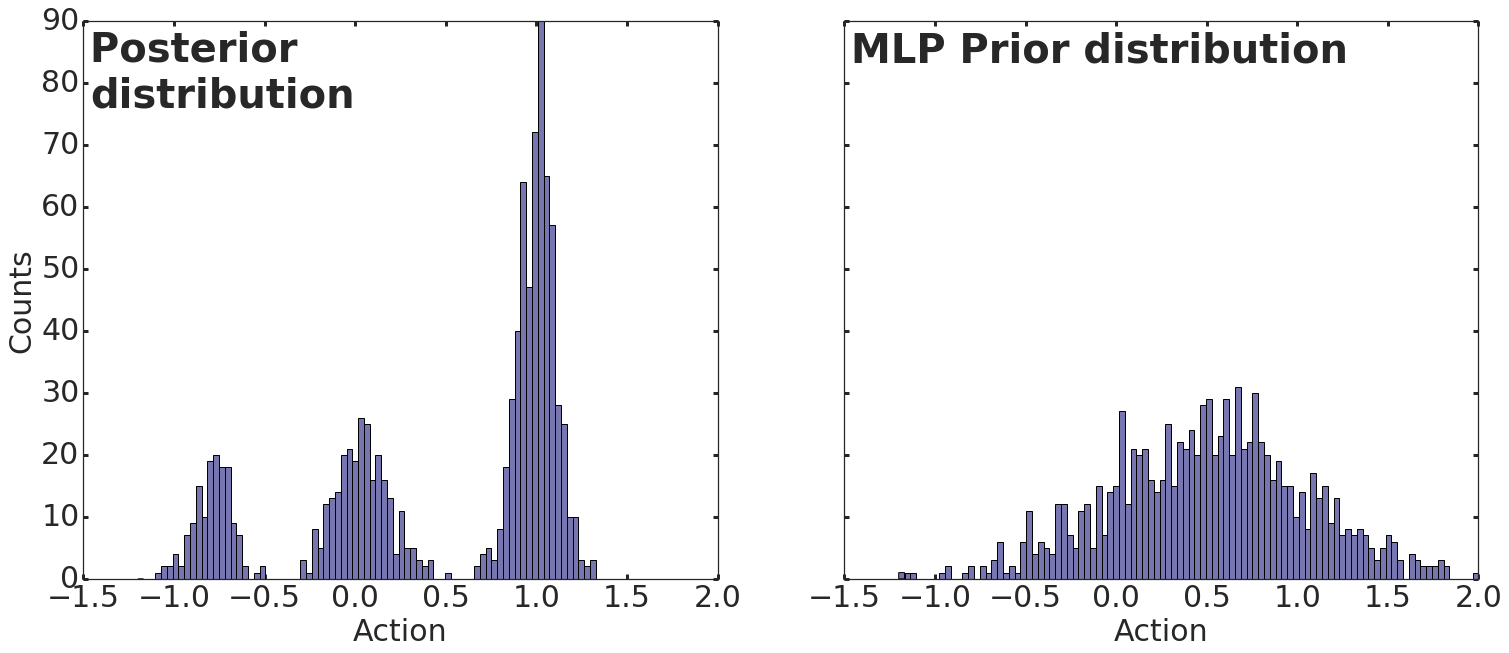} }}%
    \qquad
    \subfloat[ \label{fig:distrbn_hierarchical}]{{\includegraphics[scale=0.13]{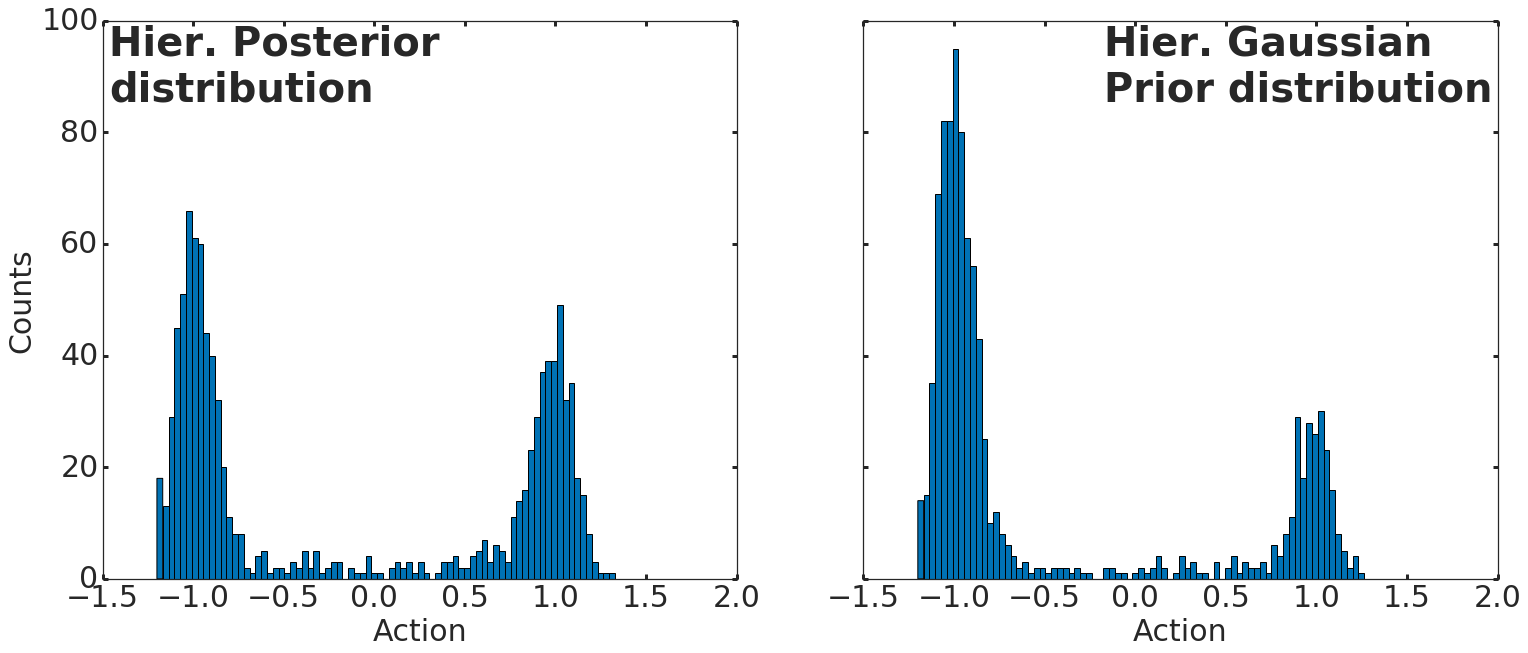} }}%
    \caption{\textbf{Empirical distributions} for different goals under the posterior and prior distributions for a) MLP and b) Hier.\ Gaussian on the \textit{Manipulation (1 box, 1 target)} task. Each plot represents the empirical distribution for an action dimension marginalized over different goals $x^G$ for a fixed goal-agnostic state $x^D$. The prior only has access to $x^D$.}
    \label{fig:distrbn_comparison}
\end{figure}

Figure \ref{fig:distrbn_comparison} helps to demonstrate this point qualitatively. In the figure, for a \emph{fixed $x^D$} (where $x^D$ is the `goal agnostic' information subset from Section \ref{sec:BP}), we sample different values of $x^G$ (the `goal specific' information) from the initial state distribution and generate samples under the posterior policy. This represents the empirical action distribution for the policy marginalized over different goals, i.e.\ the distribution that the prior is required to model. We show this distribution for a particular action dimension with the MLP model in Figure \ref{fig:distrbn_distral}. 

We observe multiple modes in the action marginal, presumably corresponding to different settings of the goal $x^G$. This marginal cannot be modeled by a unimodal Gaussian prior. We observe a mode-covering behavior where the \emph{prior} spreads probability mass across the different modes in line with the intuition described in Section \ref{sec:BP}.
In contrast, the hierarchical prior shown in Figure \ref{fig:distrbn_hierarchical} does a better job of capturing such multi-modal distributions since the action distribution $\pp^L$ is conditioned on samples from the higher level prior $\pp^H$. As we discussed at the start of Section \ref{sec:structured_models}, there may be cases where this property is desirable and may significantly affect performance.

\paragraph{Transfer:} Next, we study the effectiveness of \textit{priors} for transferring learnt behaviors. 
A policy is trained for the transfer task while regularizing against a fixed pre-trained \textit{prior}. 
We consider two transfer tasks: \textit{Manipulation (1box, 3targets)} where we transfer from a task with fewer targets ((\textit{Manipulation (1box, 1targets)}) and \textit{Manipulation (2box gather)} where we transfer from a task with a different objective (\textit{Manipulation (2box, 2targets)}). 

\begin{figure}%
    \centering
    \includegraphics[scale=0.25]{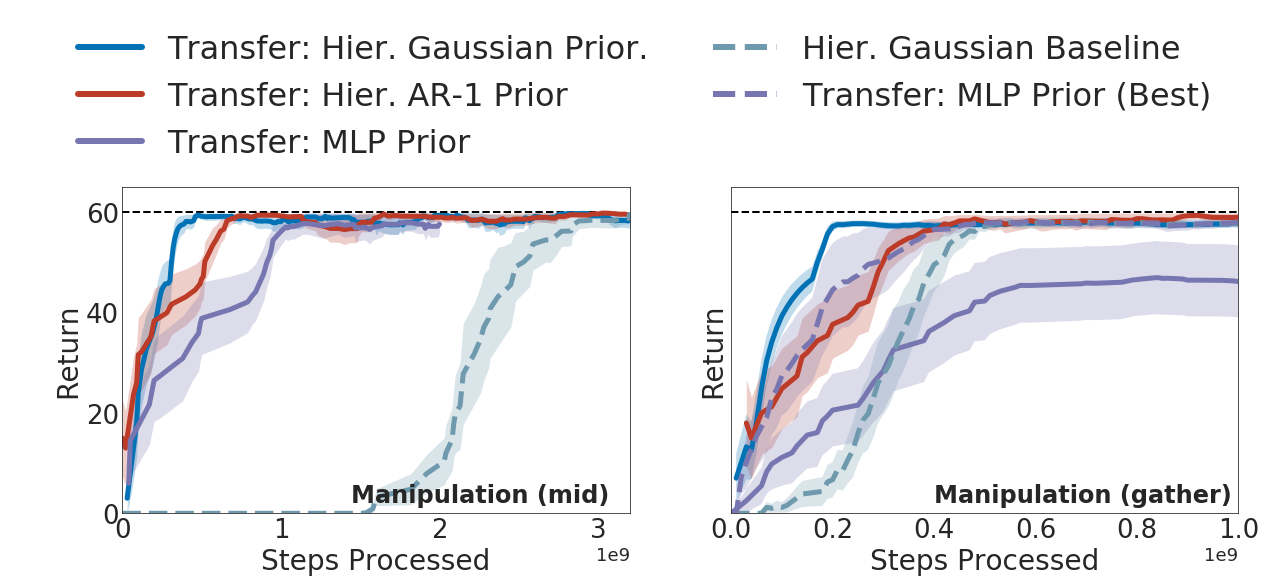}
    \caption{\textbf{Transfer performance} with various \bps on \textit{Manipulation (1b, 3t)} and \textit{Manipulation (2b, gather).}}
    \label{fig:transfer_hier}
\end{figure}

We show our results in Figure \ref{fig:transfer_hier}. We observe a significant boost in learning speed when regularizing against structured \bps across all transfer tasks. 
The advantage of hierarchical priors can be explained by the fact that the behaviors captured by the lower level policies are directly shared during transfer and do not need to be distilled through the KL term in the objective. Surprisingly, we observed that the \textit{MLP prior} slowed learning on the \textit{Manipulation (2b, gather)} task. Further analysis showed that some of the priors used for this task were of poor quality; this is visible from the performance on the training task in Figure \ref{fig:speedup_hier}c. Discarding these leads to improvement in performance shown by the dotted purple line in Figure \ref{fig:transfer_hier}.

\subsection{Separate low level modules}
All of the structured models we have considered so far used a lower level policy that was shared between the \textit{prior} $\pp$ and policy $\qp$. This can be advantageous for transfer since it allows us to directly reuse the lower level prior. 
While this is useful in many cases, it amounts to a hard constraint on the lower level which remains unchanged during transfer and cannot adapt to the demands of the target task. As an alternative, we can use a separate lower level controller for the policy as shown in Figure \ref{fig:structured_priors_descr}c. This corresponds to the full KL decomposition from Equation (\ref{eq:objective:hierarchical}). In this section, we consider an example where the separate lower level provides an advantage.

We consider a transfer task \textit{Locomotion (Gap)} where an agent must learn to \emph{jump} over a variable sized gap that is always at the same location in a straight corridor. The agent is rewarded at each time step for its forward velocity along the corridor. Locomotive behaviors from the \textit{Locomotion (Ant)} task are sufficient to jump across small gaps, but will not fare well on the larger gaps.

\begin{figure}%
    \centering
    \includegraphics[scale=0.4]{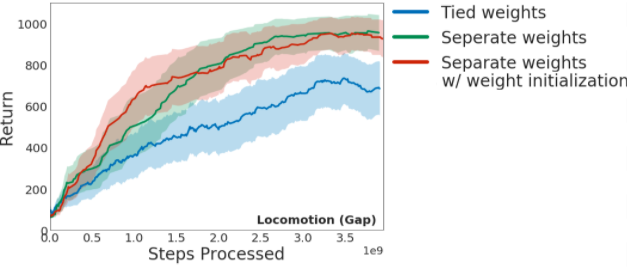}
    \caption{\textbf{Skill adaptation for hierarchical models} using separate lower level controllers on the \textit{Locomotion (Gap)} task.}
    \label{fig:transfer_hier_split}
\end{figure}

We demonstrate this point in Figure \ref{fig:transfer_hier_split}. The model with a \emph{separate} lower level performs asymptotically better than one with a shared lower level. Unlike a shared controller, the separate lower level controller adapts to the task by learning `jumping' behaviors for all gap lengths. We did not find a large gain when initializing the weights for the lower level posterior from the prior indicating that there is not much advantage of partial parameter sharing on this task.

\subsection{Sequential Navigation}
\label{sec:toy_domain_2}
In Section \ref{sec:toy_domain_1}, we discussed how \bps can model behaviors at multiple temporal scales and how modeling choices affect the nature of the behaviors learnt. Now, we will expand this analysis to the hierarchical model architectures. 

We revisit the Sequential Navigation task introduced in Section \ref{sec:toy_domain_1} which requires an agent to visit a sequence of targets in a history-dependent order. This task involves structure at various levels - goal-directed locomotion behavior and long-horizon behavior due to some target sequences being more likely to be generated than others. We consider two versions of this task with two different bodies: the 2-DoF pointmass and the 8-DoF ant walker.
Based on the intuition of Section \ref{sec:toy_domain_1}, we consider two types of models for the higher level which differ in their ability to model correlations across time: \textit{Hier.\ MLP} and \textit{Hier.\ LSTM}. For all our experiments we used an LSTM critic and an LSTM for both posterior higher level policies and MLPs for the lower level policies (full details can be found in Appendix \ref{appendix:experiment_details}). We use a similar information asymmetry to Section \ref{sec:toy_domain_1}, that is, the higher level prior and shared lower level do not have access to task-specific information. As in Section \ref{sec:toy_domain_1}, to ensure the transfer task is solvable (and not partially observable) under the MLP priors, we provide them with additional information corresponding to the index of the last visited target.

\begin{figure}%
    \centering
    \subfloat[Learning with structured priors on the easy Task\label{fig:easy_set_2}]{{\includegraphics[scale=0.25]{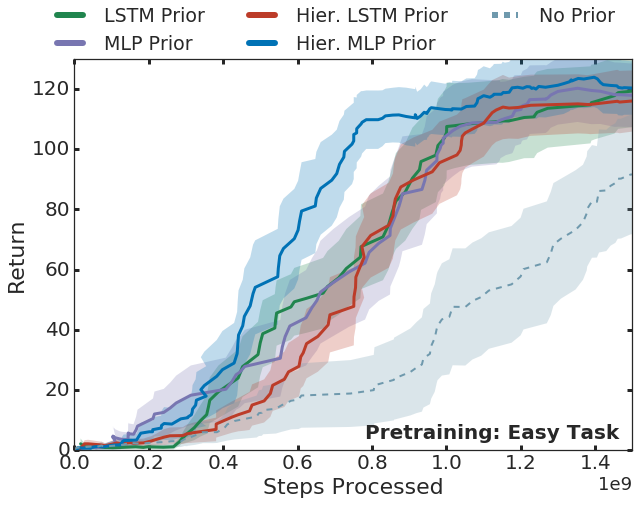} }}%
    \qquad
    \subfloat[Transfer of structured priors to hard task\label{fig:hard_set_2}]{{\includegraphics[scale=0.25]{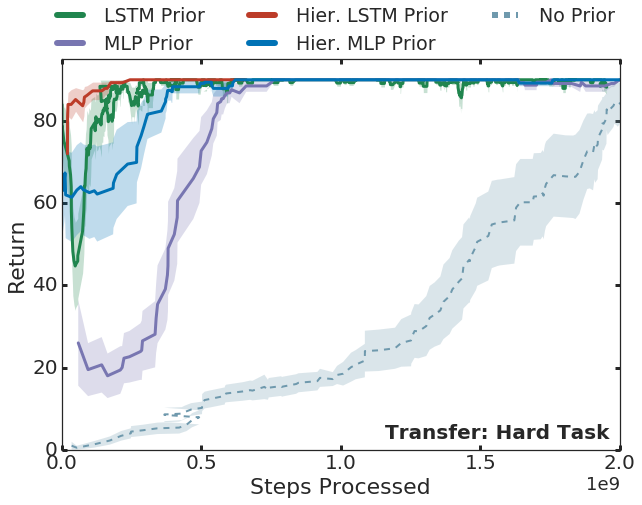} }}%
    \caption{\textbf{Learning and transfer} with structured \bps on the `Sequential Navigation' task.}
    \label{fig:pointmass_results_2}
\end{figure}

\paragraph{Training:} We first consider the effect during training in Figure \ref{fig:pointmass_results_2}. As in the previous section, we find that the structured \emph{priors} improve learning speed on this task. We also observe a slight advantage of using the \textit{Hier.\ MLP} prior in this case. This may be explained by a tradeoff between the richness of the prior and the speed at which it can be learned. 

\begin{figure}%
    \centering
    \includegraphics[scale=0.26]{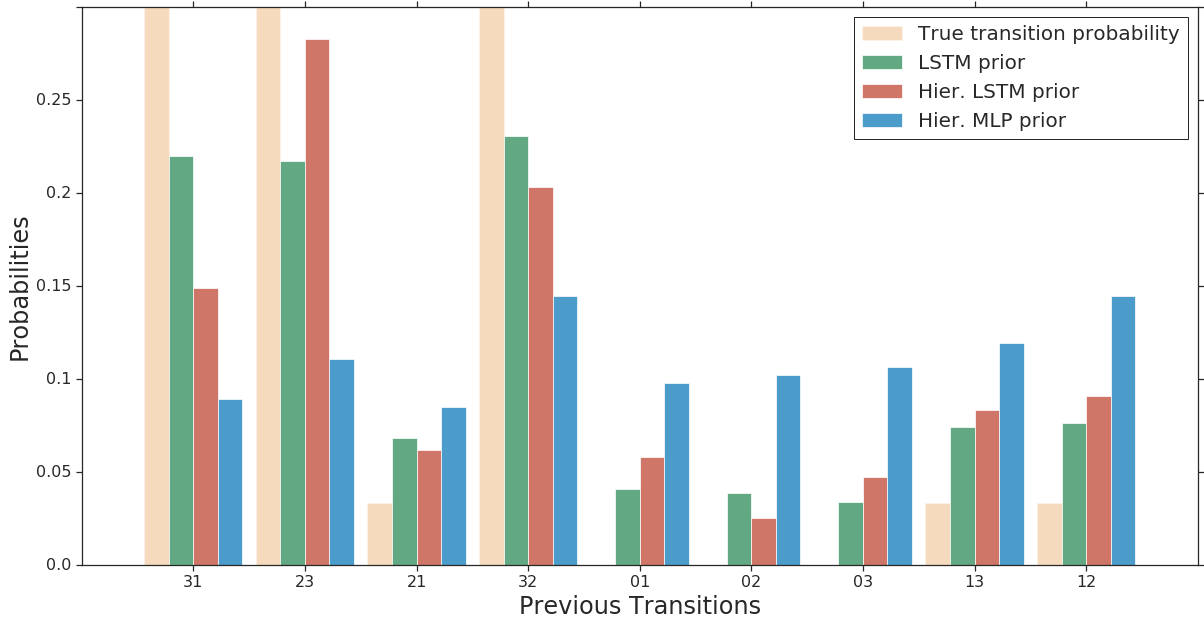}
    \caption{\textbf{Transition dynamics for first target} generated under structured \bp models. Each point on the x-axis indicates the indices of the last two targets visited. On the y-axis we plot the probability of visiting the target for each transition sequence.}
    \label{fig:transition_dynamics_2}
\end{figure}

\paragraph{Transfer:} We compare the performance of structured \emph{priors} during transfer in Figure \ref{fig:hard_set_2}. We find that the priors that model temporal structure in the task (LSTM and Hier.\ LSTM) perform better here as they are more likely to generate the rewarding sequence.
We analyze this point quantiatively in Table \ref{table:reward_rollout_kl_pointmass}. The table records the average returns generated on the transfer task over a 100 episodes when rolling out the trained priors. Priors with memory (LSTM and Hier.\ LSTM) are more likely to visit the rewarding sequence and thus generate much higher returns during early exploration. The distribution of target sequences visited  under these priors also has a lower KL divergence against the true underlying dynamics of the task which is illustrated qualitatively in Figure \ref{fig:transition_dynamics_2} and quantitatively in Table  \ref{table:reward_rollout_kl_pointmass}.
Finally, since the lower level prior (and hence policy) is fixed during transfer, there are fewer parameters to train for the hierarchical models which could explain why the structured models perform better at transfer.

\begin{table}
\centering
\begin{tabular}{ | m{4cm} | m{2cm}| m{2cm} |  m{2cm} |  m{2cm} |} 
 \hline
 \textbf{Model} & \textbf{Hier. LSTM} & \textbf{Hier. MLP} & \textbf{LSTM} & \textbf{MLP} \\
 \hline
  \hline
 
 \textbf{Return} & \textbf{81.28} & 43.82 & \textbf{82.00} & 33.94 \\ 
 \hline
 \textbf{KL} & \textbf{0.262} & 0.760 & \textbf{0.191} & 0.762 \\ 
 
 \hline

\end{tabular}
\caption{\textbf{Average return and KL}. For each pre-trained model, we compute statistics by generating trajectories under the \textit{prior} on the hard task (Averaged across a 100 episodes for 5 seeds).}
\label{table:reward_rollout_kl_pointmass}
\end{table}

\begin{figure}%
    \centering
    \subfloat[Learning with structured priors on easy task (Ant) \label{fig:ant_easy_set_2}]{{\includegraphics[scale=0.25]{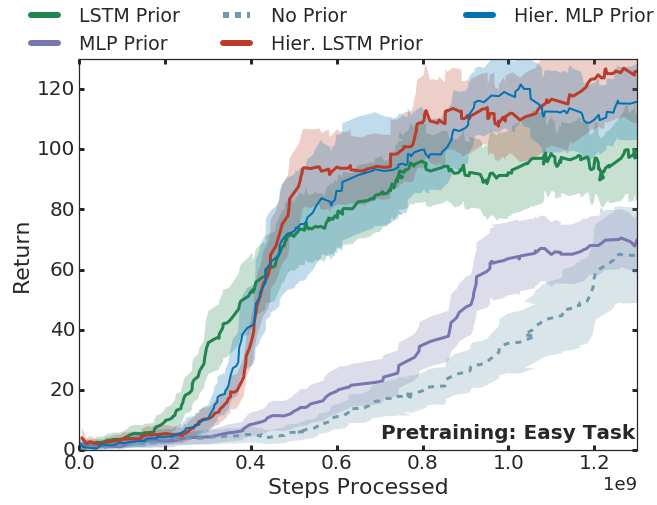} }}%
    \qquad
    \subfloat[Transfer of structured priors to hard task (Ant) \label{fig:ant_hard_set_2}]{{\includegraphics[scale=0.25]{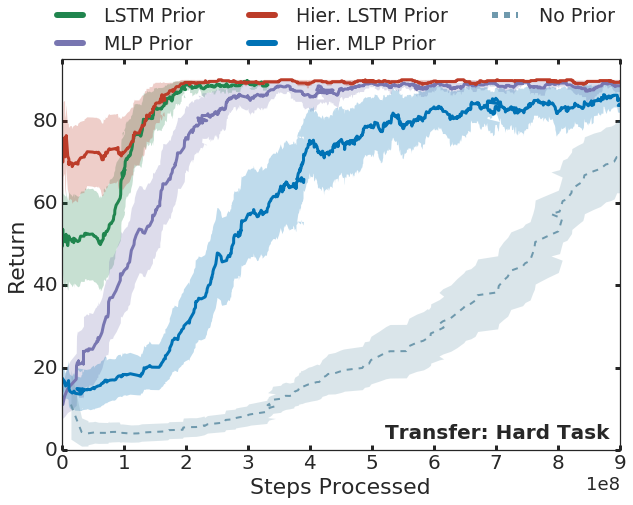} }}%
    \caption{\textbf{Learning and transfer} on \textit{Sequential Navigation (Ant)}}
    \label{fig:ant_toy_results}
\end{figure}

A possible advantage of structured models is in their ability to model different \emph{aspects} of the behavior in each layer of the hierarchy. With this in mind, we consider a version of this task with the same high-level requirements but the more complex ant body, thus posing a more difficult motor control challenge (\emph{Sequential Navigation (Ant)}).
This version of the task tests
whether a split between longer term task behaviors in the higher level and primitive movement behaviors in the lower level could prove advantageous. For this version, we restrict the targets to be generated on a smaller grid surrounding the agent in order to speed up training.
We show our results in Figure \ref{fig:ant_toy_results}. The trend that is consistent across both training and transfer is that priors with memory (LSTM and Hier.\ LSTM) learn faster than those without (MLP and Hier.\ MLP). During training, the Hier.\ MLP prior seems to learn faster than the MLP prior but the reverse holds true for transfer.
We did not find an advantage of using structured priors on this task and the reasons for this are not immediately clear. In general we found that training with parameteric latent priors (Hier. MLP and Hier. LSTM) can sometimes be unstable and care must be taken when initializing the weights for the policies (more details in Appendix 
\ref{appendix:experiment_details}). 

Overall through the results of this section and Section \ref{sec:experiments}, we have demonstrated that \bps with information and modeling constraints can improve learning speed on a range of tasks. Moreover, structured \emph{priors} can provide an additional benefit when transferring to new domains. More generally, \bps are a way to introduce inductive biases into the learning problem; a property that is likely to become increasingly important for many real world RL tasks.

\section{Discussion and Related work}
\label{sec:related}

In this section we discuss the connections between our work and various related strands of research. 
We will aim to tie the discussion to the objective introduced in Equation (\ref{eq:objective:generic_pertimestep}) and show how various algorithmic choices and modeling assumptions for $\qp$ and $\pp$ relate to established models in the literature.

\subsection{Entropy regularized RL and EM policy search}
The entropy regularized RL objective, also known as maximum entropy RL, which was considered by ~\citet{todorov2007linearly} and later by \citet{toussaint2009robot} ~\citep[also see][]{ziebart2010modeling,kappen2012optimal} has recently gained resurgence in deep reinforcement learning ~\citep[e.g.][]{fox2016taming,schulman2017equivalence,nachum2017bridging,haarnoja2017reinforcement,hausman2018learning,haarnoja2018soft}. 
As noted in Section \ref{sec:BP}, this objective is a special case of the KL regularized objective of Equation (\ref{eq:objective:generic_pertimestep}) where we set the \textit{prior} $\pp$ to be a uniform distribution over actions. This encourages the policy $\qp$ to be stochastic which is often good for exploration, has been shown to have a beneficial effect on the optimization landscape \citep{ahmed2019understanding} and may lead to more robust policies in practice \citep{haarnoja2018soft}. As discussed in Section \ref{sec:experiments}, this choice of reference distribution is general in some sense but unlikely to be good choice in all cases. 

A related family of algorithms fall under the category of \emph{expectation maximization} (EM) policy search algorithms ~\citep[e.g.][]{Peters10,toussaint2006probabilistic,rawlik2012stochastic,levine2013variational,levine2016end,montgomery2016guided,chebotar2016path,abdolmaleki2018maximum}. The fundamental idea behind this line of work is to cast policy search as an alternating optimization problem, similar to the classic EM algorithm ~\citep{dempster1977maximum}. In some cases this can be motivated from the objective in Equation  (\ref{eq:objective:generic_pertimestep}) by considering the same form for both $\qp$ and $\pp$. Then, since the form of $\pp$ is not restricted (e.g. via model capacity or information constraints), the KL term in the objective effectively acts as a trust region that constrains the optimization procedure in each step but \emph{does not} constrain the form of the final solution. This has been shown to be very successful for a large number of problems ~\citep[e.g][]{schulman2015trust,schulman2017proximal,wang2017sample,abdolmaleki2018maximum}. Yet, even though there are  algorithmic similarities and even though our results may also benefit from an effect very similar to a trust region, the motivation behind this line of work is often quite different from the ideas presented in the present paper.

The KL regularized objective has also seen application in the completely offline or Batch Reinforcement Learning (Batch RL) setting. This is often the case in applications where some prior data exists and the collection of new online data is prohibitively expensive or unsafe. Conventional off policy algorithms like DDPG typically do not fare well in this regime as demonstrated e.g.\ by \cite{fujimoto2018offpolicy, kumar2019stabilizing} since these algorithms still rely to some extent on the ability to collect new experience under the current policy. This is required to generalize beyond the data for instance, to correct for value estimation errors for previously unobserved state-action pairs. One way to mitigate this problem is to (approximately) restrict the state-action distribution to the data contained in the batch. This can be achieved by regularizing against a \bp for instance as described e.g.\ by \cite{siegel2020keep,wu2019behavior, jaques2019way, laroche2017safe, wang2020critic, peng2020advantage}.

\subsection{Information bottleneck}
\label{sec:related_info_bottleneck}
One view of the \emph{prior} touched upon in Section \ref{sec:BP} is that it encodes a `default' behavior that can be independent of some aspect of the state such as the task identity. An alternate view is that the KL-regularized objective from Equation (\ref{eq:objective:info_asym}) implements an (approximate) information bottleneck where different forms of the prior and policy penalize the information flow between the agent's interaction history and future actions in different ways.
This intuition that agents should exhibit similar behaviors in different contexts that only need to be modified or adjusted slightly e.g.~as new information becomes available, and more generally the notion that information processing in the agent is costly and subject to constraints has a significant history in the neurosciences and cognitive sciences \citep[e.g.][]{simon1956rational, wouter2018mental}. 

Formally, this idea can be expressed using an objective similar to the following:
\begin{align}
\mathcal{L_I}
&= \EE_\qp[ \sum_t \gamma^t r(s_t,a_t) - 
\alpha \gamma^t  \MutI[ x^G_t; a_t | x^D_t ]] \label{eq:objective:mi}
\end{align}
where, at time $t$, $x^G_t$ and $x^D_t$ are the goal-directed and goal-agnostic information subsets; $a_t$ is the action; $\alpha$ is a weighting factor and $\MutI$ is the conditional mutual information between $x^G_t$ and $a_t$ given $x^D_t$. This objective can be understood as attributing a cost for processing information contained in $x^G_t$ to choose action $a_t$, or, as expressing a (soft-)constraint on the capacity of the channel between $x^G_t$ and $a_t$.
As we show in Appendix \ref{appendix:mutual_info_bound}, this term can be upper bounded by:
\begin{align}
\mathcal{L_I}
&\geq \EE_\qp[ \sum_t \gamma^t r(s_t,a_t) - 
\alpha \gamma^t  \KL[ \qp(a_t | x_t) || \pp(a_t | x_t) ]] \nonumber
\end{align}
where the RHS is exactly the KL-regularized objective from Equation (\ref{eq:objective:info_asym}). Therefore, we can see our work as a particular implementation of the information bottleneck principle, where we penalize the dependence of the action on the information that is hidden from the default policy $\pp$. The models with latent variables presented in Section \ref{sec:structured_models} can be motivated in a similar way. Interestingly a similar information theoretic view of the ELBO has recently also found its way into the probabilistic modeling literature \citep[e.g.][]{alemi2016deep,alemi2017fixing}.
%

Such constraints on the channel capacity between past states and future actions have been studied in a number of works
\cite[e.g.][]{tishby2011information,ortega2011information,still2012information,rubin2012trading,ortega2013thermodynamics,tiomkin2017unified, goyal2018transfer}. 
This view also bears similarity to the formulation of \cite{strouse2018learning} where the goal is to learn to hide or reveal information for future use in multi-agent cooperation or competition. 

\subsection{Hierarchical RL}

In this work, we focus on probabilistic models of behaviors and consider hierarchical variants as a means to increase model flexibility, and to introduce modularity that facilitates partial re-use of model components. Practically, the use of hierarchical models closely resembles, for instance, that of \citealt{heess2016learning,hausman2018learning,haarnoja2018latent,merel2018neural}, in that a HL controller modulates a trained LL controller through a continuous channel and the trained LL can be subsequently reused on new tasks.

Our work, however, also emphasizes a distinction between \emph{model hierarchy} and \emph{modeling hierarchically structured} (or otherwise complex) behavior. While hierarchical models have been introduced as a way to introduce temporally consistent behaviors
\citep[e.g][]{dayan1993feudal,parr1998reinforcement,sutton1999between}, as we have seen with the LSTM priors in Sections \ref{sec:experiments} and \ref{sec:hier_experiments}, model hierarchy is not a prerequisite for instance for modeling temporally correlated, structured behavior. Furthermore, even though the modularity of our hierarchical model formulation facilitates re-use, it is not a requirement as our investigation in Section \ref{sec:experiments} demonstrates.

The perspective of \emph{hierarchical priors} has not been explored explicitly in the context of HRL. The discussion in Section \ref{sec:structured_models} focuses on a specific formulation when the latent variables present in $\pp$ and $\qp$ have the same dimensions; experimentally we explore a small set of particular models in Section \ref{sec:hier_experiments}.
Yet, the general perspective admits a much broader range of models. Latent variables in $\pp$ and $\qq$ may be continuous or discrete, and can be structured e.g. to model temporal or additional hierarchical dependencies, necessitating alternative exact or approximate inference schemes, thus giving rise to different models and algorithms. Below we illustrate how 
the KL regularized objective connects several strands of research. We present the main results here and leave the derivations to Appendix \ref{appendix:derivations}.

\paragraph{Latent variables in $\pp$} Consider the case where only $\pp$ contains latent variables. An application of Jensen's inequality leads us to the following lower bound  to $\mathcal{L}$
\begin{align}
\mathcal{L} 
&= \textstyle \EE_\qp[ \sum_t r(s_t,a_t)] - 
\KL[ \qp(\tau) || \pp(\tau) ]\\
&\geq \textstyle \EE_\qp\left [ \sum_t r(s_t,a_t) + \EE_f \left [ \log \pp(\tau|y) \right ] \right. \nonumber\\
&\hspace{1.8cm}-\left. \KL [ f(y|\tau) || \pp(y) ] \right] + \Ent [ \qp(\tau) ]. \label{eq:objective:prior_ELBO}
\end{align}
Equation (\ref{eq:objective:prior_ELBO}) is an application of the evidence lower bound to $\log \pp(\tau)$ and $f(y| \tau)$ is the approximate posterior, just like, for instance, in variational auto-encoders (VAE; \citealt{kingma2013autoencoding, rezende2014stochastic}). This emphasizes the fact that $\pp$ can be seen as a model of the trajectory data generated by $\qp$. Note however, as explained in Section \ref{sec:BP} the model extends only to the history-conditional action distribution (the policy) and not the system dynamics.
If $y$ is discrete and takes on a small number of values exact inference may be possible. If it is conveniently factored over time, and discrete, similar to a hidden Markov model (HMM), then we can compute $f(y|\tau)$ exactly (e.g. using the forward-backward algorithm in  \citep{rabiner1989tutorial} for HMMs). For temporally factored continuous $y$ a learned  parameterized approximation to the true posterior \citep[e.g.][]{chung2015recurrent}, or mixed inference schemes \citep[e.g.][]{johnson2016composing} may be appropriate.

Variants of this formulation have been considered, for instance, in the imitation learning literature where the authors model trajectories generated by one or multiple experts using a parameteric decoder $\pp(\tau| y)$ and encoder $f(y| \tau)$. In this view $\qp$ is fixed (it may correspond to one or more RL policies, or, for instance, a set of trajectories generated by a human operator) and we optimize the objective only with respect to $\pp$.
\citet{fox2017multi,krishnan2017discovery} learn a sequential discrete latent variable model of trajectories  (an \emph{option model}) with exact inference.
\citet{merel2018neural} and \citet{wang2017robust} model trajectories generated by one or multiple experts using a parameteric encoder $f(y| \tau)$ and decoder $\pp(\tau| y)$.
The resulting models can then be used to solve new  tasks by training a higher level controller to modulate the learnt decoder $\pp(\tau| y)$ similar to our transfer experiments in Section \ref{sec:hier_experiments} \citep[e.g.][]{merel2018neural}.

\paragraph{Latent variables in $\qp$} 
On the other hand, if only $\qp$ contains latent variables $z_t$ this results in the following approximation that can be derived from Equation (\ref{eq:objective:info_asym}) (see Appendix \ref{appendix:posterior_bound_derivation} for proof):
\begin{align}
\mathcal{L} &\geq 
\EE_\qp[ \sum_t r(s_t, a_t) ] 
+ \EE_\qp \left [ \log \pp(\tau) + \log g(z | \tau) \right ] + \Ent [ \qp(\tau | Z) ] + \Ent [\qp(Z) ]
\label{eq:objective:posterior_bound_full}
\end{align}
where $g$ is a learned approximation to the true posterior $\qq(z|\tau)$. 
This formulation is also discussed in \cite{hausman2018learning}.
It is important to note that the role of $g$ here is only superficially similar to that of a variational distribution in the ELBO since the expectation is still taken with respect to $\qq$ and \emph{not} $g$. However, the optimal $g$, which makes the bound tight, is still given by the true posterior $\qp(z| \tau)$. The effect of $g$ here is to exert pressure on $\qp(\tau| z)$ to make (future) trajectories that result from different $z$s to be `distinguishable' under $g$. 

\cite{hausman2018learning} considers the special case when $z$ is sampled once at the beginning of the episode. In this case, $g$ effectively introduces an intrinsic reward that encourages $z$ to be identifiable from the trajectory, and thus encourages a solution where different $z$ lead to different outcomes. This formulation has similarities with diversity inducing regularization schemes based on mutual information \citep[e.g.][]{gregor2016variational,florensa2017stochastic,eysenbach2018diversity}, but arises here as an approximation to trajectory entropy. More generally this is related to approaches based on empowerment \citep{klyubin2005empowerment, salge2013empowerment, mohamed2015variational} where mutual information is used to construct an intrinsic reward that encourages the agent to visit highly entropic states.

This formulation also has interesting connections to auxiliary variable formulations in the approximate inference literature \citep{salimans2014bridging,agakov2004an}. A related approach is that of \cite{haarnoja2018latent}, who consider a particular parametric form for policies with latent variables for which the entropy term can be computed analytically and no approximation is needed.

From an RL perspective, $g$ can be thought of as a learned \emph{internal reward} which is dependent on the choice of $z$ i.e. we can think of it as a goal-conditioned internal reward where the internal goal is specified by $z$. As noted above, since the expectation in Equation (\ref{eq:objective:posterior_bound_full}) only considers samples from $\qp$, we are free to condition $g$ on subsets of the information set, albeit at the expense of loosening the bound.
We can use this as another way of injecting prior knowledge by choosing a particular form of $g$. For instance, it is easy to see that for $z,s \in \mathbb{R}^D$ and $g(z|s) = N(z | s, \sigma^2)$ we can effectively force $z$ to play the role of a goal state, and $g$ to compute an internal reward that is proportional to $|| z-s||^2$. This view draws an interesting connection between the auxiliary variable variational formulations to other `subgoal' based formulations in the RL literature \citep[e.g.][]{dayan1993feudal,vezhnevets2017feudal,nachum2018data}.

The options framework proposed by \cite{sutton1999between, precup2000temporal} and more recently studied e.g. by \cite{bacon2016optioncritic,fox2017multi, frans2018meta} introduces discrete set of (sub-)policies (`options') that are selected by a (high-level) policy and then executed until an option-specific termination criterion is met (at which point a new option is chosen for execution). Options thus model recurrent temporal correlations between actions and can help improve exploration and credit assignment.

From a probabilistic perspective, options are a special case of our framework where a discrete latent variable or option $c_t$ is held constant until some termination criterion is satisfied. The model consists of three components: a binary switching variable $b_t$ which is drawn from a Bernoulli distribution $\beta(. | x_t, c_{t-1})$ that models option termination i.e. whether to continue with the current option or switch to a new one; a high level distribution over options $\qp(c_t | x_t)$ and an option specific policy $\qp_{c_t}(a_t| x_t)$. Together with these components, we have the following model:
\begin{align}
\qp(a_t, z_t | x_t, z_{t-1}) &= \sum_{b \in \{0, 1\}} \beta(b_t| x_t, c_{t-1}) \left [ \qp(c_t| x_t)^{b_t} + \delta(c_t, c_{t-1})^{1-b_t}\right ] \qp_{c_t}(a_t | x_t) \nonumber \\
&= \beta(0| x_t, c_{t-1}) \delta(c_t, c_{t-1}) \qp_{c_t}(a_t | x_t) + 
\beta(1| x_t, c_{t-1}) \qp(c_t | x_t)  \qp_{c_t}(a_t | x_t) \nonumber
\end{align}
\\
where $c_t \in \{ 1 \dots K \}$ indexes the current option and $z_t=(c_t, b_t)$. The presence of the time-correlated latent variable allows the model to capture temporal correlations that are not mediated by the state. A similar probabilistic model for options has also been considered by \citet{daniel2016hierarchical, daniel2016probabilistic} and more recently by \citet{wulfmeier2020dataefficient} although they do not consider a notion of a \emph{prior} in those cases.

One important question in the options literature and more generally in hierarchical reinforcement learning is how useful lower level policies or options can be learned \citep{barto2002, Schmidhuber91neuralsequence, weiring1997hq, dietterich1999hierarchical, boutilier1997prioritized}. From a probabilistic modeling perspective options are beneficial when a domain exhibits temporal correlations between actions that cannot be expressed in terms of the option policies' observations. This can be achieved, for instance by introducing information asymmetry into the model, and a suitable prior can further shape the learned representation (and achieve e.g.\ effects similar to \cite{harb2017waiting} where a `deliberation cost' for switching between options can be introduced through an information bottleneck as discussed in Section \ref{sec:structured_models} in the form of a \emph{prior} on the behavior of the option variable $b_t$). Our results have also shown, however, that latent variables (and options in particular) are not necessary to model such temporal correlations. They can be similarly captured by unstructured auto-regressive models and then be used, for instance, for exploration (e.g.\ Sections \ref{sec:toy_domain_1} and \ref{sec:toy_domain_2}). This observation is in keeping with the results of \citet{riedmiller2018learning} who consider a setting where data is shared among a set of related tasks that vary in difficulty. The skills learnt by solving the easier tasks results in improved exploration on the hard tasks. They demonstrate that this exploration benefit is enough to solve challenging real world tasks even with non-hierarchical policies. Similarly, \citep{nachum2019does} empirically demonstrated that the advantages of sub-goal based \citep{nachum2018data, levy2017learning} and options based \citep{bacon2016optioncritic, precup2000temporal} HRL approaches can largely be attributed to better exploration. 
Finally structured representations learnt by hierarchical option policies do hold additional computational (rather than representational) benefits, which could be useful for planning \citep[e.g.][]{sutton1999between, precup2000temporal, krishnan2017discovery}. However this is not the focus on the present work.

\section{Conclusion}
\label{sec:conclusion}

We can summarize the findings of our work as follows:
\begin{itemize}
    \item \textbf{\textit{Behavioral Priors} can speed up learning}: We have demonstrated a method to learn \bps, which are policies that capture a general space of solutions to a number of related tasks.
    We have shown that using these \emph{priors} during learning can lead to significant speedups on complex tasks. We have shown that \bps can model long term correlations in action space like movement patterns for complex bodies or complex task structure in the case of `Sequential Navigation' tasks. While we have only considered one simple algorithmic application of this idea by extending the SVG-0 algorithm, the methods introduced here can easily be incorporated into any RL learning objective.
    \item \textbf{Information regularization can lead to generalization}: We have explored how restricting processing capacity through information asymmetry or modeling capacity through architectural constraints can be used as a form of regularization to learn general behaviors. We have demonstrated that this idea can be extended to structured models to train modular hierarchial polices. In contrast to other methods, in our approach, this behavior emerges during learning without explicitly building it into the task or architecture.    
    \item \textbf{Structured Priors can be advantageous}: We have shown how models that incorporate structure in the form of latent variables can model more complex distributions; a feature that can be important in some domains.
    \item \textbf{General framework for HRL}: We discussed how our formulation can be related to a larger set of ideas within RL. The latent variable formulation presented in Section \ref{sec:structured_models} can be connected to a number of ideas in HRL and more generally we have shown the relation between the KL-regularized RL objective and objectives based on mutual information and curiosity.
\end{itemize}

Ultimately though, all of the methods that we have discussed in this work draw from a rich, mature literature on probabilistic modeling. Most of the ideas here largely stem from a simple principle: policies and task solutions define distributions over a space of trajectories. This perspective, which comes quite naturally from the KL-regularized objective, allows us to reason about \emph{priors} which cover a broader manifold of trajectories. While we have considered a specific form of this where the \emph{prior} is learnt jointly in a multi-task scenario; more generally the \emph{prior} introduces a means to introduce inductive biases into the RL problem. We have considered this from a modeling perspective through the use of a hierarchical, latent variable prior but the potential applications extend well beyond this. 

For instance in applications like robotics, simulation is often expensive and we may only have access to some other form of data. In other cases where safety is a factor, exploration often needs to be limited to space of `safe' trajectories that may be learnt or hand defined in some way. As RL methods gain wider application to real world domains, these needs are bound to become increasingly important. The real world is often challenging to work with, impossible to perfectly simulate and quite often constrained in the kinds of solutions that can be applied. In such settings, the ability to introduce prior knowledge about the problem in order to constrain the solution space or for better exploration is likely to be of importance. In that regard, we think that the methods presented here can help with these issues. 

However this is only a starting point; a doorway to new algorithms that couple the fields of reinforcement learning with probabilistic modeling. We continue to be excited by the avenues created by this marriage and hope it will continue to bear fruit with new ideas and research in the years to come.

{\noindent \em Remainder omitted in this sample. See http://www.jmlr.org/papers/ for full paper.}


\acks{We would like to acknowledge Gregory Wayne and Siddhant Jayakumar
for immensely useful discussion. We would also like to thank Josh Merel and Tom Schaul for their valuable comments and guidance on earlier drafts of this work. We would also like to acknowledge David Szepesvari for their input on the role of hierarchy in RL. A large part of this work is the fruit of these discussions. Finally, we are indebted to the valuable contributions of Courtney Antrobus for her support, wisdom and organizational wizardry without which this work would have taken far longer to produce.}


\newpage
\appendix
\section{Probabilistic modeling and variational inference}
\label{appendix:probablistic_background}

In this section, we briefly describe core ideas from probabilistic machine learning and their connection with the KL-regularized objective as described in the text. We refer the reader to \cite{bishop2006pattern, koller2009probabilistic, murphy2012machine} for introductory textbooks on probabilistic learning. 

Probabilistic learning is framed in the context of learning a generative model $\model$ which is a joint distribution of some data  $\data_\all = \{ \data_1, \dots \data_N \}$, parameterised by $\param$, potentially involving latent variables  $\latent$, and  including contextual information or covariates.

For simplicity, though the framework is more general, we will assume that the data is \emph{independent and identically distributed (iid)} under the model, i.e.\, 
\begin{align}
\model(\data_\all) = \prod_{i=1}^N \model(\data_i)
= \prod_{i=1}^N \int \model(\data_i|\latent_i)\model(\latent_i) d\latent_i
\end{align}
where we have also introduced one iid latent variable $\latent_i$ corresponding to each observed datum $\data_i$. For intuition, each datum can be an image while the latent variable describes the unobserved properties of the visual scene. $\model(z)$ is a prior distribution, e.g.\ over visual scenes, and $\model(x|z)$ is an observation model, e.g.\ of the image given the scene.

In this framework, \emph{inference} refers to computation or approximation of the posterior distribution over latent variables given observed data given by Bayes' rule, i.e.,\ 
\begin{align}
\model(\latent|\data)
= \frac{\model(\data|\latent)\model(\latent)}{\model(\data)}
= \frac{\model(\data|\latent)\model(\latent)}{\int \model(\data|\latent)\model(\latent)d\latent}
\end{align}
while \emph{learning} refers to estimation of the model parameters $\param$ from data. Typically this is achieved by maximising the log \emph{likelihood}, i.e.\ the log marginal probability of the data as a function of the parameters $\param$:
\begin{align}
    \param^\text{MLE} = \arg\max_\param \log \model(\data_\all)
    = \arg\max_\param \sum_{i=1}^N \log \model(\data_i) \label{eq:mle}
\end{align} 
In this paper we will only be interested in maximum likelihood learning. It is possible to extend the framework to one of Bayesian learning, i.e.\ to treat the parameters themselves as random variables, whose posterior distribution $p(\param|\data)$ captures the knowledge gained from observing data $\data_\all$.

The core difficulty of the probabilistic learning framework is in the computation of the log marginal and posterior probabilities (as well as their gradients).  These are intractable to compute exactly for most models of interest. 

A common approach is to employ a variational approximation. This stems from the following lower bound on the log marginal probability obtained by applying Jensen's inequality. For any conditional distribution $\varpost(\latent|\data)$ and any $x$, we have:
\begin{align}
    \log \model(\data) &= \log \int \model(\data, \latent) d\latent = \log \int \varpost(\latent|\data)\frac{\model(\data, \latent)}{\varpost(\latent|\data)} d \latent \label{eq:log-marginal}\\
    &= \log \EE_\varpost \left [ \frac{\model(\data, \latent)}{\varpost(\latent|\data)} \right ]
    \geq \EE_\varpost\left[ \log \frac{\model(\data,\latent)}{\varpost(\latent|\data)} \right]
    \label{eq:elbo-1}\\
    &= \EE_\varpost\left[\log \model(\data|\latent) + \log\frac{\model(\latent)}{\varpost(\latent|\data)}\right]  
    =  \EE_\varpost\left[\log \model(\data|\latent)\right] - \KL [ \varpost(\latent|\data) || \model(\latent) ]
    \label{eq:elbo-2}\\
    &= \EE_\varpost\left [  \log \model(\data) + \log \frac{\model(\latent|\data)}{\varpost(\latent|\data)} \right ] 
    =  \log \model(\data) - \KL [ \varpost(\latent|\data) || \model(\latent | \data) ],
    \label{eq:elbo-3}
\end{align}
where we use Jensen's inequality for lower bound in \eqref{eq:elbo-1}.  \eqref{eq:elbo-2} and \eqref{eq:elbo-3} show two equivalent ways of writing the resulting bound, the first of which is tractable and the second of which is intractable but gives a different way to see that the expression gives a lower bound on the log marginal probability, since the KL divergence is non-negative.

This bound is often referred to as evidence lower bound or \emph{ELBO}. As the last line shows the bound is tight when $\varpost$ matches the true posterior $\model(\latent | \data)$. For most models the true posterior is as intractable as the log marginal probability\footnote{This is not surprising considering that $\model(\latent | \data) = \model(\data,\latent) / \model(\data)$, i.e.\ the normalization constant of the posterior is exactly the intractable marginal probability.} and $\varpost$ is thus chosen to belong to some restricted family parameterised by variational parameters $\varparam$.

Plugging this lower bound into \eqref{eq:mle}, we get the following objective to be maximized with respect to model parameters $\param$ and variational parameters $\varparam$:
\begin{align}
    \sum_{i=1}^N \log \model(\data_i)
    \ge \sum_{i=1}^N
    \EE_{\varpost_\varparam(\latent_i|\data_i)}\left[
    \log\model(\data_i|\latent_i) + \log\model(\latent_i) - \log\varpost_\varparam(\latent_i|\data_i)\right] \label{eq:elbo:objective}
\end{align}

\section{2-D Task Results}
\label{appendix:additional_results}

In this section we present additional results to help build an intuition for the joint optimization scheme of the policy and \bp from Equation (\ref{eq:objective:info_asym}). We consider a 2-D setting where each task consists of a continuous probability distribution function (pdf) and the reward for a given 2-D action corresponds to the log-likelihood under the distribution. For most of our tasks we use a multivariate Gaussian distribution for which the reward for each task is maximal at the mean. We also consider one special task whose underlying distribution is a mixture of two Gaussians.

\begin{figure}%
    \centering
    \includegraphics[width=\linewidth]{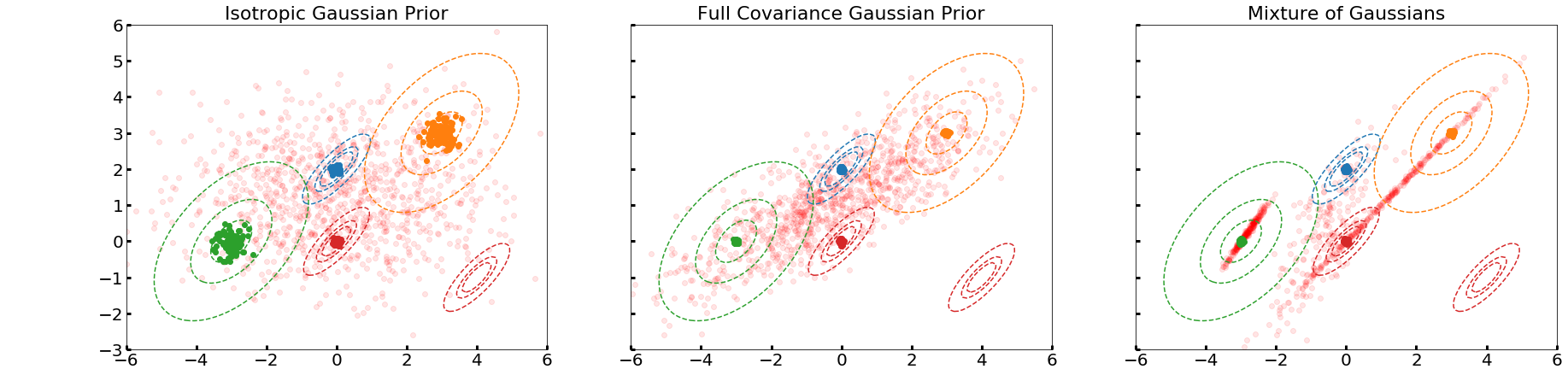}
    \caption{\textbf{Training on 2D task.} Each colored dotted contour represents a reward function for the different tasks. Each figure shows samples from the task specific posteriors (colored dots) and \bps (red) for different models.}
    \label{fig:toy_pretrain}
\end{figure}

We consider a multi-task setup consisting of 4 different tasks shown as colored contours in Figure \ref{fig:toy_pretrain}. The distributions for each task are: (green) a gaussian with mean (-3., 0.), scale (0.5, 0.5) and correlation $\rho$ 0.5; (blue) a gaussian with mean (0, 2.), scale (0.2, 0.2) and correlation $\rho$ 0.8; (orange) a gaussian with mean (3., 3.), scale (0.5, 0.5) and correlation $\rho$ 0.5; (red) a mixture of two equally likely gaussians with mean (0, 0.) and (-4, -1) and an equal of scale (0.2, 0.2) with correlation 0.8. Each plot shows task specific posteriors (colored dots) and \emph{priors} (red dots) for different prior models. We consider three models which, in increasing order of expressivity are: isotropic Gaussian; Gaussian with full covariance and a mixture of Gaussians. All posteriors are multivariate Gaussians.

In Figure \ref{fig:toy_pretrain} we observe that the task specific posteriors generate samples close to the center of the reward contours (where the reward is maximal). For the mixture task, the posteriors learn to model the solution closer to the distribution modeled under the \emph{prior} which is exactly as expected from the $\KL$ in Equation (\ref{eq:objective:generic_pertimestep}). The \emph{priors} capture a more general distribution containing \emph{all} the task solutions. Moreover, the choice of \emph{prior} affects the kinds of solutions learnt. The least expressive model - an isotropic Gaussian (Figure \ref{fig:toy_pretrain}a), cannot capture the joint solution to all tasks perfectly.
In contrast, a mixture of Gaussians \emph{prior} (Figure \ref{fig:toy_pretrain}c) almost perfectly models the mixture distribution of the posteriors. The joint optimization of Equation (\ref{eq:ProbabilisticRL:MultitaskKL}) has an interesting effect on the posterior solutions. More general \emph{priors} lead to noisy posteriors as in Figure \ref{fig:toy_pretrain}a) since the objective trades off task reward against minimizing the KL against the \emph{prior}.

\begin{figure}%
    \centering
    \includegraphics[width=\linewidth]{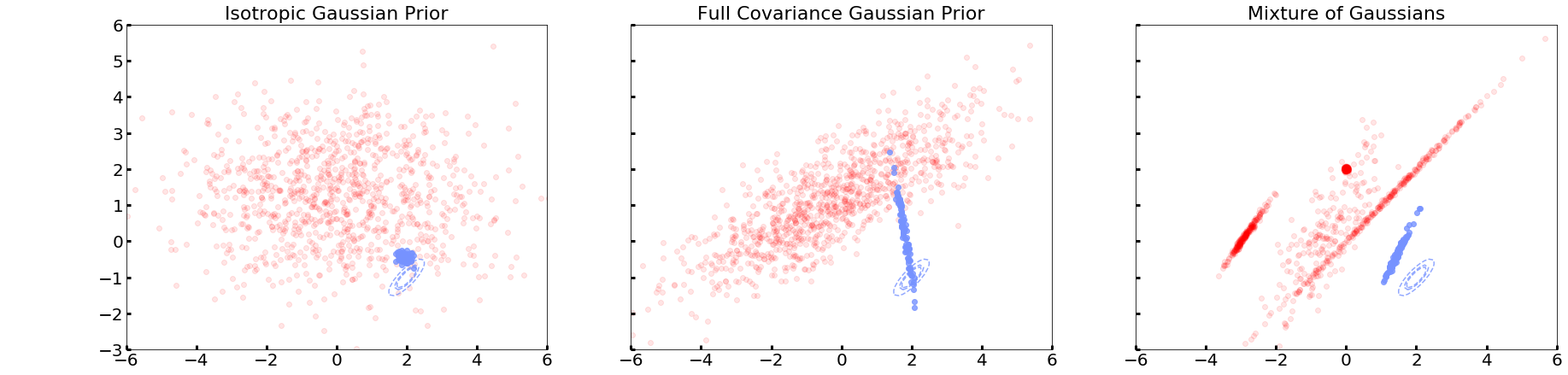}
    \caption{\textbf{Transfer to new 2D task.} Each dotted contour represents a reward function with each color representing a different tasks. Each figure plots samples from task specific posteriors (colored) and \bps (red) for various \emph{prior} models. Posteriors were only allowed to train for 10 iterations.}
    \label{fig:toy_transfer}
\end{figure}

In Figure \ref{fig:toy_transfer} we consider the effect of \emph{priors} on a held-out transfer task. We consider a new task defined by a gaussian distribution centered at (2, -1) with a scale of (0.1, 0.1) and a correlation of 0.8. We consider a few-shot transfer setting where policies are trained for only 10 iterations using the objective from Equation (\ref{eq:objective:generic_pertimestep}). This highlights the advantage of a more general \emph{prior}.The more general trajectory space captured by the isotropic Gaussian is better suited for a transfer task as illustrated in Figure \ref{fig:toy_transfer}. In contrast the mixture of Gaussians \emph{prior} from Figure \ref{fig:toy_transfer}c does not fare as well.
\section{Derivations}
\label{appendix:derivations}

Here, we present derivations for the lower bounds and proofs presented in the main text.

\subsection{Decomposition of KL per-timestep}
\label{appendix:per_timestep_KL}
In this part we demonstrate that the KL over trajectory distributions decomposes conveniently to a sum over per timestep KLs. We start with the general KL formulation over trajectories:
\begin{align}
\mathcal{L}
&= \EE_\qp[ \sum_{t=0}^\infty \gamma^t r(s_t,a_t)] - \KL[ \qp(\tau) || \pp(\tau) ] \nonumber
\end{align}

If we choose $\pp$ and $\qp$ to be distributions over trajectories that arise as a composition of the true system dynamics $p(s_{t+1} | s_t, a_t)$ and state-conditional distributions over actions ($\pp(a_t | s_t)$ and $\qp(a_t | s_t)$) we obtain the following factorization:
\begin{align}
\pp(\tau) &= p(s_0) \prod_t \pp(a_t | s_t) p(s_{t+1} | s_t, a_t) \label{eq:objective:p_simple}\\
\qp(\tau) &= p(s_0) \prod_t \qp(a_t | s_t) p(s_{t+1} |s_t,a_t), \label{eq:objective:q_simple}
\end{align}

in which case we recover exactly the expected reward objective in Section \ref{sec:BP}, with a per-step regularization term that penalizes deviation between $\pp$ and $\qp$:
\begin{align}
\mathcal{L}(\pp,\qp) &= \EE_\qp[ \sum_t (\gamma^t r(s_t,a_t) + \gamma^t \log \frac{\pp(a_t | s_t)}{\qp(a_t | s_t ) } )] \nonumber \\
&= \EE_\qp[ \sum_t \gamma^t (r(s_t,a_t) - 
\KL[ \qp || \pp] )]. \nonumber
\end{align}

\subsection{Bound when \texorpdfstring{$\pp$}{} contains latent}
\label{appendix:prior_elbo_derivation}
In this part, we derive the bound presented in Equation (\ref{eq:objective:prior_ELBO}). In this setting we consider the case where $\pp$ contains latent variables $y$. In that case, we have:
\begin{align}
\mathcal{L}
&= \textstyle \EE_\qp[ \sum_t \gamma^t (r(s_t,a_t)] - 
\gamma^t \KL[ \qp(\tau) || \pp(\tau) ] \nonumber \nonumber\\
&\geq \textstyle \EE_\qp \left [ \sum_t \gamma^t \left ( r(s_t,a_t) + \EE_f \left [ \log\frac{\pp(\tau,y)}{f(y|\tau)} \right ] \right ) \right] + \gamma^t \Ent [ \qp(\tau) ] \nonumber\\
&= \textstyle \EE_\qp\left [ \sum_t \gamma^t (r(s_t,a_t) + \EE_f \left [ \log \pp(\tau|y) \right ]) \right. \nonumber\\
&\hspace{1.8cm}-\left. \gamma^t \KL [ f(y|\tau) || \pp(y) ] \right] + \gamma^t \Ent [ \qp(\tau) ].
\end{align}

\subsection{Bound when \texorpdfstring{$\qp$}{} contains latent}
\label{appendix:posterior_bound_derivation}
In what follows we expand on the derivation for the bound introduced in Equation (\ref{eq:objective:posterior_bound_full}). We begin with the objective for $\mathcal{L}$ from Equation (\ref{eq:objective:generic_pertimestep}) and consider the case where only $\qp$ contains latent variables $z_t$ which could be continuous. This results in the approximation as discussed in \cite{hausman2018learning}:
\begin{align}
\mathcal{L} &= \EE_\qp[\sum_t \gamma^t r(x_t, a_t) ] 
- \gamma^t \EE_\qp [ \KL[ \qp(\tau) || \pp(\tau) ] ] \nonumber \\
&\geq 
\EE_\qp[\sum_t \gamma^t r(x_t, a_t) ] 
+ \gamma^t \EE_\qp  \left [ \log \frac{\pp(\tau)g(z | \tau)}{\qp(\tau | Z)\qp(Z) } \right ] \nonumber
\\
&= \EE_\qp[\sum_t \gamma^t r(x_t, a_t) ] 
+ \gamma^t \EE_\qp  \left [ \log \pp(\tau) + \log g(z | \tau) \right ] + \gamma^t \Ent [ \qp(\tau | Z) ] + \gamma^t \Ent [\qp(Z) ]
\end{align}

\subsection{Bound when both \texorpdfstring{$\pp$}{} and \texorpdfstring{$\qp$}{} contain latents}
\label{appendix:hierarchical_objective_derivation}
Here we present the proof for the main result presented in Section \ref{sec:structured_models}. 
\begin{align}
\KL[\qp(a_t|x_t) || \pp(a_t|x_t)]
\le& \KL[\qp(a_t|x_t) || \pp(a_t|x_t)] + \EE_{\qp(a_t|x_t)}[\KL[\qp^H(z_t|x_t) || \pp^H(z_t|x_t)]] \nonumber \\
&= \EE_{\qp(a_t|x_t)}[
\textstyle \log \frac{\qp(a_t|x_t)}{\pp(a_t|x_t)}] \nonumber\\
 &\hspace{1.8cm}\textstyle +\EE_{\qp(a_t|x_t)}[
\EE_{\qp^H(z_t|a_t,x_t)}[\textstyle \log \frac{\qp^H(z_t|a_t,x_t)}{\pp^H(z_t|a_t,x_t)}]
\nonumber \\
&= \EE_{\qp(a_t,z_t|x_t)}[
\textstyle \log \frac{\qp(a_t,z_t|x_t)}{\pp(a_t,z_t|x_t)}]
=
\KL(z_t,a_t|x_t) \nonumber \\
&= \EE_{\qp^H(z_t|x_t)}[
\textstyle \log \frac{\qp^H(z_t|x_t)}{\pp^H(z_t|x_t)}] \nonumber \\
 &\hspace{1.8cm}\textstyle + \EE_{\qp^H(z_t|x_t)}[
\EE_{\qp^L(a_t|z_t,x_t)}[\textstyle \log \frac{\qp^L(a_t|z_t,x_t)}{\pp^L(a_t|z_t,x_t)}]
\nonumber \\
&= \KL[\qp^H(z_t|x_t) || \pp^H(z_t|x_t)] \nonumber \\
&\hspace{1.8cm}\textstyle + \EE_{\qp^H(z_t|x_t)}[\KL[\qp^L(a_t|z_t, x_t) || \pp^L(a_t|z_t, x_t)] ] \label{eq:kl_bound_proof}
\end{align}

\subsection{Upper Bounding Mutual Information}
\label{appendix:mutual_info_bound}

We begin with the objective for mutual information that was introduced in Section \ref{sec:related}:
\begin{align}
\mathcal{L_I}
&= \EE_\qp[ \sum_t \gamma^t r(s_t,a_t) - 
\alpha \gamma^t  \MutI[ x^G_t; a_t | x^D_t ]] \nonumber
\end{align}

The mutual information term can be upper bounded as follows:
\begin{align}
\MutI[x^G_t; a_t | x^D_t] &= \int \qp(x^G_t, a_t| x^D_t) \log \frac{\qp(x^G_t, a_t| 
x^D_t)}{\qp(x^G_t| x^D_t)\qp(a_t| x^D_t)} \nonumber \\
 &= \int \qp(x^G_t| x^D_t) \qp(a_t| x^G_t, x^D_t) \log \frac{\qp(x^G_t| x^D_t)  \qp(a_t| x^G_t, x^D_t)}{\qp(x^G_t| x^D_t)\qp(a_t| x^D_t)} \nonumber \\
 &= \int \qp(x^G_t| x^D_t) \qp(a_t| x_t) \log \frac{\qp(a_t| x_t)}{\qp(a_t| x^D_t)} \nonumber \\
 &\leq \int \qp(x^G_t| x^D_t) \qp(a_t| x_t) \log \frac{\qp(a_t| x_t)}{\pp(a_t| x^D_t)} \nonumber \\
 &= \EE_\qp[ \KL[ \qp(a_t| x_t) || \pp(a_t| x^D_t) ]] \label{eq:mi_kl_proof}
\end{align}

since 
\begin{align}
\KL[ \qp(a_t| x_t) || \pp(a_t| x^D_t) ] &\geq 0 \nonumber \\
\EE_\qp[ \log \frac{\qp(a_t | x_t)}{\pp(a_t | x^D_t)} ] &\geq 0 \nonumber \\
\EE_\qp[\log \qp(a_t | x_t)] &\geq \EE_\qp[\log \pp(a_t | x^D_t)] \nonumber
\end{align}

We can extend the bound in Equation (\ref{eq:mi_kl_proof}) to the hierarchical setting by combining it with the bound from Equation (\ref{eq:kl_bound_proof}). In the case when the LL is shared between $\qp$ and $\pp$ (and thus the lower KL is exactly 0), this reduces to:
\begin{align}
\MutI[x^G_t; a_t | x^D_t]  &\leq \KL[\qp^H(z_t|x_t) || \pp^H(z_t|x_t^D)] \nonumber
\end{align}
\section{Environments}
\label{appendix:environments}

In this section, we describe the detailed configuration of the continuous control tasks and bodies used. Videos depicting these tasks can be found at the accompanying website:  \url{https://sites.google.com/view/behavior-priors}.

\subsection{Tasks}
\label{appendix:tasks}
\paragraph{Toy Domain: PointMass on Grid}
The $1\times1$ arena consists of the agent at the center and 4 targets placed uniformly at random on the square. On each episode a sequence of targets is generated according to 2nd order Markovian dynamics based on the previous 2 targets visited based on the distribution given in Table \ref{table:appendix_pointmass_dynamics}. Note that the dynamics are chosen in such a way such that the marginal probabilities are uniformly distributed when conditioned on a single target. We consider two versions of the task: \textit{easy task} where the agent is given the next target to visit and the sequence continues until the episode terminates (after 375 steps). For the \textit{hard task}, the agent must visit the 4 targets in order. The agent receives a reward of +10 for every correct target and an additional bonus of +50 for completing the \textit{hard task}. For the \textit{easy task} we generate sequences with $p=0.9$ while for the hard task we generate sequences with $p=1.0$ (refer to Table \ref{table:appendix_pointmass_dynamics}). The policy  and \textit{priors} receive the location of all the targets and the agent's global position and velocity. Additionally, for the \textit{easy task}, the policy gets the next target in sequence while for the  \textit{hard task} it receives the 6-length sequence as well as the index of the next target. 

\begin{table}[htp]
\centering
\begin{tabular}{ | m{2cm} | m{2cm} | m{2cm} | m{2cm} | m{2cm} |} 
 \hline
 \textbf{Sequence} & \textbf{Target 0} & \textbf{Target 1} & \textbf{Target 2} & \textbf{Target 3} \\
 \hline
 \hline
 
      \textbf{10} & 0.00 & 0.00 & $p$ & 1-$p$ \\
      \hline
      \textbf{20} & 0.00 &  1-$p$ &  0.00 &  $p$ \\
      \hline
      \textbf{30} & 0.00 &  $p$ &  1-$p$ &  0.00 \\
      \hline

      \textbf{01} & 0.00 &  0.00 &  $p$ &  1-$p$ \\ 
      \hline
      \textbf{21} & 1-$p$ &  0.00 &  0.00 &  $p$ \\
      \hline
      \textbf{31} & $p$ &  0.00 &  1-$p$ &  0.00 \\ 
      \hline

      \textbf{12}: & 1-$p$ &  0.00 &  0.00 &  $p$ \\ 
      \hline
      \textbf{32}: & $p$ &  1-$p$ &  0.00 &  0.00 \\ 
      \hline
      \textbf{02}: & 0.00 &  $p$ &  0.00 &  1-$p$ \\ 
      \hline

      \textbf{03}: & 0.00 &  $p$ &  1-$p$ &  0.00 \\
      \hline
      \textbf{13}: & 1-$p$ &  0.00 &  $p$ &  0.00 \\ 
      \hline
      \textbf{23}: & $p$ & 1-$p$ &  0.00 &  0.00 \\ 
      \hline
      
\end{tabular}
\caption{\textbf{Markov Transition Dynamics} used to generate the target sequence for the point mass task. $p$ indicates the probability of visiting that target and the row indices indicate the previous two targets visited.}
\label{table:appendix_pointmass_dynamics}
\end{table}

\paragraph{Locomotion (Humanoid)}
We use the open-source version of the \href{https://github.com/deepmind/dm_control/blob/master/dm\_control/locomotion/tasks/go\_to\_target.py}{go\_to\_target} task from the DeepMind control suite \citep{tassa2018control}. We use an arena of $8\times8$
with a moving target and a humanoid walker. The humanoid is spawned in the center of the arena and a target is spawned uniformly at random on the square on each episode. The agent receives a reward of 1 when is within a distance of 1 unit from the target for up to 10 steps. At the end of this period, a new target is spawned uniformly at random in a $1.5\times1.5$ area around the walker. New targets continue to be spawned in this manner for the duration of the episode (400 steps). The agent receives proprioceptive information and the egocentric location of the target relative to its root position.

\paragraph{Locomotion (Ant)}
On a fixed $8\times8$ arena, an ant walker and 3 targets are placed at a random location and orientation at the beginning of each episode.
In each episode, one of the 3 targets is chosen uniformly at random to be the \textit{goal}. The agent receives a reward of +60 if its root is within a $0.5\times0.5$ area around the target. The episode terminates when the agent reaches the target or after 400 timesteps. The agent receives proprioceptive information, the location of all the targets relative to its current location and a onehot encoding of the \textit{goal} target index as input.


\paragraph{Locomotion and Manipulation}
On a fixed $3\times3$ arena, an ant walker, 2 targets and a cubic box of edge $0.5$ are placed at a random location and orientation at the beginning of each episode.
In each episode, one of the 2 targets is chosen uniformly at random to be the \textit{agent goal} and the other becomes the \textit{box goal}. There are 2 components to this task: the agent receives a reward of +10 if its root is within a $0.5\times0.5$ area around the \textit{agent goal} and a +10 if the center of the box is within \textit{box goal}. If the agent completes \textit{both} components of the tasks, it receives a bonus reward of +50. The episode terminates when the agent and box are at their goals or after 400 timesteps. The agent receives proprioceptive information, the location of all the targets relative to its current location, the location of the box relative to its current location and a onehot encoding of the \textit{agent goal}.
 
\paragraph{Manipulation (1 box, $k$ targets)}
On a fixed $3\times3$ arena, an ant walker, $k$ targets and a cubic box of edge $0.5$ are placed at a random location and orientation at the beginning of each episode.
In each episode, one of the $k$ targets is chosen uniformly at random to be the \textit{box goal}. The agent receives a reward of +60 if the center of the box is within \textit{box goal}. The episode terminates when box is at the goal or after 400 timesteps. The agent receives proprioceptive information, the location of all the targets relative to its current location, the location of the box relative to its current location and a onehot encoding of the \textit{box goal}. In our experiments we have two tasks where $k=1$ and $k=3$.

\paragraph{Manipulation (2 box, $k$ targets)}
On a fixed $3\times3$ arena, a ball walker, $k$ targets and a cubic box of edge $0.5$ are placed at a random location and orientation at the beginning of each episode.
In each episode, one of the $k$ targets is chosen uniformly at random to be the \textit{first box goal} and another is chosen to be the \textit{second box goal} . The agent receives a reward of +10 if the center of the first box is within the \textit{first box goal} or the center of the second box is within the \textit{second box goal}. If both boxes are at their goals, the agent gets a bonus reward of +50. The episode terminates when both boxes are at their goals or after 400 timesteps. The agent receives proprioceptive information, the location of all the targets relative to its current location, the location of the boxes relative to its current location and a onehot encoding of the \textit{first box goal}. In our experiments we have $k=2$.
 
\paragraph{Manipulation (2 box gather)}
This is a variation of the Manipulation (2 box, 2 targets) as described above where the agent receives a +60 for bring both boxes together such that their edges touch. The episode terminates when both boxes touch or after 400 timesteps.

\subsection{Bodies}
\label{appendix:bodies}

We use three different bodies: Ball, Ant, and Humanoid as defined by the DeepMind control suite \citep{tassa2018control}.
These have been all been used in several previous works \citep{heess2017emergence,xie2018transferring,galashov2018information}
The \textbf{Ball} is a body with 2 actuators for moving forward or backward, turning left, and turning right.
The \textbf{Ant} is a body with 4 legs and 8 actuators, which moves its legs to walk and to interact with objects.
The \textbf{Humanoid} is a body with a torso and 2 legs and arms and 23 actuators.
The \textit{proprioceptive} information provided by each body includes the body height, position of the end effectors, the positions and velocities of its joints and sensor readings from an accelerometer, gyroscope and velocimeter attached to its torso. 
\section{Experiment details}
\label{appendix:experiment_details}

Throughout the experiments, we use 32 actors to collect trajectories and a single learner to optimize the model. For the transfer experiments in Section \ref{sec:toy_domain_1} and Section \ref{sec:toy_domain_2} however we use 4 actors per learner. We plot average episode return with respect to the number of steps processed by the learner.
Note that the number of steps is different from the number of agent's interaction with environment. By steps, we refer to the batches of experience data that are samples by a centralized learner to update model parameters.
When learning from scratch we report results as the mean and standard deviation of average returns for 5 random seeds.
For the transfer learning experiments, we use 5 seeds for the initial training, and then two random seeds per model on the transfer task. Thus, in total, 10 different runs are used to estimate mean and standard deviations of the average returns. Hyperparameters, including KL cost and action entropy regularization cost, are optimized on a per-task basis. More details are provided below.

\subsection{Hyperparameters used for experiments}
\label{appendix:hyper_parameters}

For most of our experiments we used MLP or LSTM torso networks with an final MLP layer whose output dimensionality is twice the action space for the policy and \textit{prior}. The output of this network is used as the mean and the log scale to parameterize an isotropic Gaussian output. For stability and in keeping with prior work \citep[e.g.][]{heess2015learning}, we pass the log scale output through a softplus layer and add an additional fixed noise of 0.01 to prevent collapse. The hierarchical policies described in Section \ref{sec:hier_experiments} use a similar approach where the output of the higher level layer is used to paramterize a Gaussian from which we sample the latent $z$. We also use target networks for actor, \textit{prior} and critic for increased stability.

Below we provide the default hyperparameters used across tasks followed by modifications made for specific tasks as well as the range of parameters swept over for each task. All MLP networks used \textit{ELU} activations. The output of the Gaussian actions was capped at 1.0.

\subsubsection{Default parameters}

\textit{Actor learning rate:} $\beta_{pi}$ = 1e-4. 
\textit{Critic learning rate:} $\beta_{pi}$ = 1e-4.
\textit{Prior learning rate:} $\beta_{pi}$ = 1e-4.

\textit{Target network update period:} = 100
\textit{HL policy Torso:} MLP with sizes (200, 10)
\textit{LL policy Torso:} MLP with sizes (200, 100)
\textit{Critic Torso:} MLP with sizes (400, 300).

\textit{Batch size:} 512
\textit{Unroll length:} 10

\textit{Entropy bonus:} $\lambda$ = 1e-4.
\textit{Distillation cost:} $\alpha$ =  1e-3.
\textit{Posterior entropy cost:} $\alpha$ =  1e-3.
\textit{Number of actors:} 32

\subsubsection{Task specific parameters}
For each task we list any parameters that were changed from the default ones and specify sweeps within square brackets ([]).

\paragraph{PointMass: Easy task}

\textit{MLP/LSTM Actor torso:} (64, 64)
\textit{MLP/LSTM Prior torso:} (64, 64)
\textit{Critic torso:} LSTM with size (128, 64)

\textit{Entropy bonus:} [1e-4, 0.0]
\textit{Distillation cost:} [0.0, 1e-1, 1e-2, 1e-3]
\textit{Unroll length:} 40

\textit{Actor/Critic/Prior learning rate:} [5e-5, 1e-4, 5e-4]

\textit{Hier. MLP/Hier. LSTM Prior HL torso:} [ (64, 4),  (64, 10),  (64, 100)]
\textit{Hier. MLP/Hier. LSTM Posterior HL torso:} [ (64, 4),  (64, 10),  (64, 100)]
\textit{Hier. MLP/Hier. LSTM LL torso:} [ (64, 64)]

\paragraph{PointMass: Hard task}

\textit{MLP/LSTM Actor torso:} (64, 64)
\textit{MLP/LSTM Prior torso:} (64, 64)
\textit{Critic torso:} LSTM with size (128, 64)

\textit{Entropy bonus:} [1e-4, 0.0]
\textit{Distillation cost:} [0.0, 1e-1, 1e-2, 1e-3]
\textit{Unroll length:} 40

\textit{Actor/Critic/Prior learning rate:} [5e-5, 1e-4, 5e-4]

\textit{Hier. MLP/Hier. LSTM Prior HL torso:} [ (64, 4),  (64, 10),  (64, 100)]
\textit{Hier. MLP/Hier. LSTM Posterior HL torso:} [ (64, 4),  (64, 10),  (64, 100)]
\textit{Hier. MLP/Hier. LSTM LL torso:} [ (64, 64)]
\textit{Number of actors:} 4

\paragraph{Locomotion (Humanoid)}

\textit{MLP Actor torso:} (200, 100)
\textit{MLP Prior torso:} (200, 100)
\textit{Distillation cost:} [1e-1, 1e-2, 1e-3, 1e-4]

\textit{Actor learning rate:} [5e-5, 1e-4, 5e-4]
\textit{Critic learning rate:} [5e-5, 1e-4, 5e-4]
\textit{Prior learning rate:} [5e-5, 1e-4, 5e-4]
\textit{Target network update period:} [50, 100]

\paragraph{Locomotion (Ant)}

\textit{MLP Actor torso:} (200, 100)
\textit{MLP Prior torso:} (200, 100)
\textit{HL Policy torso:} [(200, 4), (200, 10), (200, 100)]

\textit{Distillation cost:} [1e-1, 1e-2, 1e-3, 1e-4]
\textit{Actor/Critic/Prior learning rate:} [5e-5, 1e-4, 5e-4]

\paragraph{Locomotion and Manipulation}

\textit{MLP Actor torso:} (200, 100)
\textit{Encoder for proprioception:} (200, 100)
\textit{Encoder for target:} (50)
\textit{Encoder for box:} (50)
\textit{Encoder for task encoding:} (50)
\textit{HL Policy torso:} [(100, 4), (100, 10), (10), (100)]

\textit{Distillation cost:} [1e-1, 1e-2, 1e-3]
\textit{HL Policy torso:} [(200, 4), (200, 10), (200, 100)]


\paragraph{Manipulation (1 box, 1 target)}

\textit{HL Policy torso:} [(200, 4), (200, 10), (200, 100)]
\textit{LL Policy torso:} [(200, 4), (200, 10), (200, 100)]
\textit{Policy/MLP Prior Torso:} [(200, 4, 200, 100), (200, 10, 200, 100), (200, 100, 200, 100)]

\textit{Parameter for AR-1 process:} [0.9, 0.95]

\textit{Distillation cost:} [1e-2, 1e-3, 1e-4]
\textit{Actor/Critic/Prior learning rate:} [1e-4, 5e-4]

\paragraph{Manipulation (1 box, 1 target)}

\textit{Distillation cost:} [1e-2, 1e-3, 1e-4]
\textit{HL Policy torso:} [(200, 4), (200, 10), (200, 100)]
\textit{Policy/MLP Prior Torso:} [(200, 4, 200, 100), (200, 10, 200, 100), (200, 100, 200, 100)]

\textit{Parameter for AR-1 process:} [0.9, 0.95]

\textit{Actor/Critic/Prior learning rate:} [1e-4, 5e-4]

\paragraph{Manipulation (1 box, 3 targets)}

\textit{Distillation cost:} [1e-2, 1e-3, 1e-4]
\textit{HL Policy torso:} [(200, 4), (200, 10), (200, 100)]
\textit{Policy/MLP Prior Torso:} [(200, 4, 200, 100), (200, 10, 200, 100), (200, 100, 200, 100)]

\textit{Actor/Critic/Prior learning rate:} [1e-4, 5e-4]
\textit{Parameter for AR-1 process:} [0.9, 0.95]

\paragraph{Manipulation (2 box, 2 targets)}
\textit{Distillation cost:} [1e-2, 1e-3, 1e-4]
\textit{HL Policy torso:} [(200, 4), (200, 10), (200, 100)]
\textit{Policy/MLP Prior Torso:} [(200, 4, 200, 100), (200, 10, 200, 100), (200, 100, 200, 100)]

\textit{Actor/Critic/Prior learning rate:} [1e-4, 5e-4]
\textit{Parameter for AR-1 process:} [0.9, 0.95]

\paragraph{Manipulation (2 box gather)}
\textit{Distillation cost:} [1e-2, 1e-3, 1e-4]
\textit{HL Policy torso:} [(200, 4), (200, 10), (200, 100)]
\textit{Policy/MLP Prior Torso:} [(200, 4, 200, 100), (200, 10, 200, 100), (200, 100, 200, 100)]

\textit{Actor/Critic/Prior learning rate:} [1e-4, 5e-4]
\textit{Parameter for AR-1 process:} [0.9, 0.95]








\vskip 0.2in
\bibliography{ref}

\end{document}